\documentclass[10pt,a4paper,logo]{googledeepmind}

\usepackage{natbib}
\usepackage{subcaption}
\usepackage{multirow}
\usepackage{array}
\usepackage{tikz}

\title{OnePred: Next-Query Prediction via Recursive Intent Memory in Multi-Turn Conversations}

\author[1,*]{Jiangwang Chen}
\author[1,*]{Bowen Zhang}
\author[1,*]{Zixin Song}
\author[2,*]{Jiazheng Kang}
\author[2]{Xiao Yang}
\author[2]{Da Zhu}
\author[2]{Guanjun Jiang}
\affil[1]{Tsinghua University\\
\texttt{\{jw-chen24,zbw23,songzx24\}@mails.tsinghua.edu.cn}}
\affil[2]{Qwen Business Unit\\
\texttt{\{kangjiazheng.kjz,yx501135,zhuda.zd,guanj.jianggj\}@alibaba-inc.com}}
\renewcommand{\authornotetext}{\textsuperscript{*}Equal contribution.}

\begin{abstract}
Although large language model (LLM) conversational systems process millions of multi-turn dialogues daily, they remain fundamentally reactive: they respond only after the user types a query. A key step toward proactive interaction is next-query prediction, which anticipates the user's subsequent query based solely on the preceding dialogue. Progress on this task is hindered by the lack of dedicated benchmarks and a fundamental efficiency--quality trade-off: naively concatenating full dialogue history incurs linearly growing token consumption, while truncating to the latest turn discards crucial cross-turn context. Our key insight is that accurate prediction does not require re-reading raw history; it suffices to track the user's evolving intent trajectory across topics, unresolved needs, and interest shifts. We propose OnePred, which maintains a recursively updated memory as its sole cross-turn context, bounding the per-turn cost independently of conversation length. We train the model via a two-stage reinforcement learning pipeline that first teaches what to predict, then what to compress, shaping the memory into a prediction-oriented intent chain. To establish a rigorous testbed, we introduce NQP-Bench, spanning three diverse subsets. Experiments demonstrate that OnePred reduces per-turn token consumption by up to 22$\times$ compared to full-history inputs while consistently exceeding all baselines in prediction quality, with larger gains on longer conversations.
\end{abstract}

\begin{document}
\maketitle

\section{Introduction}
\label{sec:intro}


LLM-based conversational systems now process millions of multi-turn dialogues daily~\citep{zheng2024lmsys}, yet they remain fundamentally \emph{reactive}: each response waits for the user to type. This passivity imposes real costs. Users must re-articulate needs that the system could have anticipated, retrieval pipelines cannot pre-fetch relevant documents, and latency accumulates over successive exchanges. We study \emph{next-query prediction}, the task of forecasting a user's next query from the preceding dialogue, which enables the system to act before the user explicitly speaks.


Next-query prediction can shift conversational systems from reactive to proactive, enabling various downstream applications. The system can suggest follow-up questions to help users articulate unformulated needs, analogous to ``People Also Ask'' in web search. Predicted queries
also enable speculative execution, allowing the system to pre-fetch documents, invoke retrieval-augmented generation pipelines, or compute candidate answers before the user submits, greatly reducing perceived latency. In addition, predicted queries allow routing infrastructure to dispatch anticipated requests to specialized models or knowledge bases ahead of time, improving throughput and response quality. More broadly, accurately predicting the next query suggests that the system has captured the user's evolving intent, a capability with direct implications for long-term user experience.


Despite this practical value, next-query prediction has received little dedicated study. Progress is impeded by both data scarcity and quality: authentic multi-turn logs are limited, and many real-world conversations are not naturally predictable, requiring careful curation to identify genuinely predictable instances. Related problems in adjacent fields do not fill this gap. \emph{Query suggestion} in conversational systems typically optimizes for user clicks via preference alignment on implicit human feedback~\citep{min2025ctr, yin2026clicks}, rather than modeling the natural trajectory of user intents. \emph{Proactive dialogue} steers conversations toward specific goals~\citep{deng2025proactive,deng2023survey,wu2019proactive} but focuses on system actions rather than anticipating user behavior.


Two straightforward approaches to handling dialogue history each fall short. A \emph{Current-turn} model observes only the latest exchange. While efficient, it cannot detect topic continuations or recurring needs that span earlier turns, limiting its predictive capability in extended conversations. In contrast, a \emph{Full-history} model concatenates all previous turns, providing rich context for prediction but at an increasing cost. As dialogue lengthens, the required context window grows linearly, demanding expensive long-context infrastructure or forcing lossy truncation. Moreover, predictive signals are buried under lengthy assistant responses and topical digressions, creating a signal-to-noise problem that simple concatenation cannot resolve. The core challenge is therefore one of compression: effective prediction requires cross-turn context, but only the subset that is informative for the user's next query. 


%


Our key insight is that a model need not load the entire raw history into its context window. Instead, it only needs to track the user's intent chain: the evolving trajectory across topics, unresolved needs, and interest shifts. We represent this chain as a bounded free-form text state, allowing the LLM to read and write memory natively without additional modules while keeping the content directly interpretable. We propose \textbf{OnePred}, in which this memory is recursively updated at each turn. The model receives only the previous memory and the current user--assistant exchange, and decides what to retain, update, or discard. This design addresses both limitations above: it preserves cross-turn context that a single-turn window lacks, while bounding per-turn inference cost and filtering predictive signals from irrelevant context.


Learning to maintain such memory is non-trivial. Since there is no ground-truth annotation for what the memory should contain, its quality can only be evaluated through downstream prediction performance, making standard supervised fine-tuning (SFT) unsuitable. At the same time, optimizing memory poses a cold-start problem: effective compression requires knowing what signals matter for prediction, yet learning to predict through memory requires the memory to already carry useful context. To break this circular dependency, we adopt a two-stage reinforcement learning (RL) pipeline.
\textbf{Stage~1} (Full-History RL) gives the model the complete conversation and trains next-query prediction directly.
\textbf{Stage~2} (Agentic Memory RL) removes history access and trains the model to predict through its memory alone, forcing it to learn what to preserve.
The ordering is essential: a model cannot learn to build useful memory without first knowing what information serves prediction.

To the best of our knowledge, no dedicated benchmark currently exists for open-ended next-query prediction in multi-turn LLM-assistant conversations, making it difficult to measure progress or compare approaches. 
To bridge this gap, we construct \textbf{NQP-Bench} (\S\ref{sec:benchmark}).

Our contributions are threefold:
\begin{itemize}[itemsep=0pt,topsep=2pt]
\item \textbf{Benchmark.} We construct \textbf{NQP-Bench}, to our knowledge the first dedicated benchmark for open-ended next-query prediction in multi-turn LLM-assistant conversations, spanning private, public, and cross-source settings, paired with a graded LLM-judge evaluation protocol validated against human annotators.
\item \textbf{Method.} We propose OnePred, which maintains a recursive memory as a prediction-oriented intent chain. It is trained via a two-stage RL pipeline: Stage~1 learns to predict from full history, and Stage~2 learns to predict through bounded memory alone, thereby discovering what context should be preserved.
\item \textbf{Empirical Findings.} OnePred reduces per-turn token consumption by up to 22$\times$ compared with full-history inputs while exceeding all baselines in prediction quality across datasets and model configurations, with its advantage increasing on longer conversations.
\end{itemize}

\section{NQP-Bench}
\label{sec:benchmark}
Given a multi-turn user--assistant conversation $\mathcal{C}_T = \{(q_t, r_t)\}_{t=1}^{T}$, the task of \emph{next-query prediction} is to predict the user's next query $q_{T+1}$ from the conversation context.
To rigorously evaluate this capability, we require access to authentic multi-turn conversations where users interact naturally with LLM assistants. Consequently, we source data from real deployment logs and two large-scale public corpora, treating the last user query $q_{T+1}$ in each session as the ground-truth prediction target and all preceding turns $\mathcal{C}_T$ as context.
From these sessions we construct \textbf{NQP-Bench} (Next-Query Prediction Benchmark), comprising three subsets:
\textbf{NQP-Priv}, drawn from a commercial conversational AI deployment and reflecting authentic user behavior;
\textbf{NQP-Wild}, derived from WildChat~\citep{zhao2024wildchat}\footnote{\url{https://huggingface.co/datasets/allenai/WildChat-4.8M}} to support reproducibility;
and \textbf{NQP-Share}, derived from ShareChat~\citep{sharechat-2025}\footnote{\url{https://huggingface.co/datasets/tucnguyen/ShareChat}} for cross-source generalization evaluation. NQP-Bench targets context-grounded next-query prediction rather than unconstrained future-behavior forecasting. To strictly protect user privacy, NQP-Priv will remain closed-source. The NQP-Wild and NQP-Share subsets are publicly released to support community research.

\begin{figure}[t]
    \centering
    \includegraphics[width=0.80\linewidth]{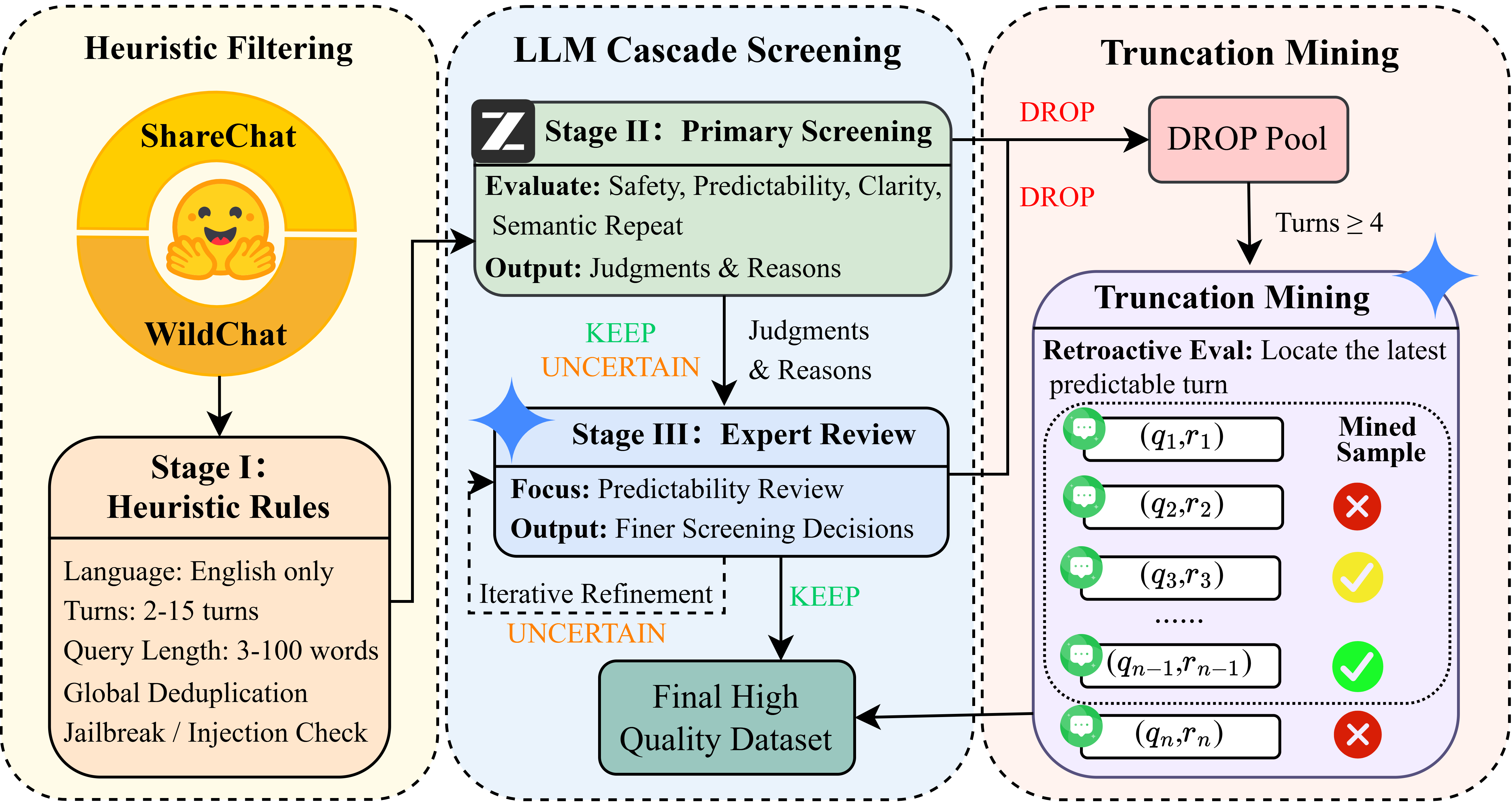}
    \caption{Overview of the NQP-Bench construction pipeline. The process integrates heuristic filtering, a two-stage LLM cascade for rigorous predictability screening, and a retroactive truncation mining strategy to salvage high-quality conversation prefixes from the DROP pool.}
    \label{fig:dataset_construction}
\end{figure}

\paragraph{Dataset Construction.}
All three subsets undergo a unified multi-stage curation pipeline illustrated in Figure~\ref{fig:dataset_construction}.
Stage~I applies heuristic rules to filter out low-quality noise. 
It enforces basic boundaries on turn counts and query lengths, removes degenerate patterns such as jailbreak attempts, and executes global deduplication for the public corpora.

Following heuristic filtering, an LLM cascade screening strategy spanning Stage~II and Stage~III is deployed to assess target predictability.
Stage~II serves as the primary screening phase using GLM-5~\citep{zeng2026glm} to evaluate safety, predictability, target clarity, and semantic repetition. 
The primary quality criterion is predictability: the core intent and key information of the target query must be inferable and learnable from the conversation history. 
To enforce this criterion, the model evaluates each sample against strict unpredictability rules, flagging cases such as random topic jumps or the sudden injection of user-private information. 
Based on this evaluation, each sample is categorized as KEEP, UNCERTAIN, or DROP, accompanied by a textual justification.
Subsequently, Stage~III employs a stronger expert language model, Gemini 2.5 Pro \citep{comanici2025gemini}, to review all KEEP and UNCERTAIN samples. 
The expert model reads the initial judgment and justification, evaluates whether the reasoning is valid, and decides whether to confirm or overturn the verdict.
For boundary cases initially marked as UNCERTAIN, the expert model performs an iterative refinement process to resolve ambiguities and reach a final decision.

To increase data yield from complex dialogues, we apply a truncation mining strategy to samples assigned to the DROP pool. For rejected conversations with four or more turns, we retroactively examine the dialogue history to identify the latest predictable turn. We then truncate the dialogue at this boundary, recovering high-quality predictable prefixes instead of discarding the entire session.

All retained samples are further annotated with three fine-grained labels (intent categories, cognitive difficulty, and intent transfer paradigms), with attrition rates and final statistics provided in Appendix~\ref{app:attrition} and Table~\ref{tab:dataset_stats}. Averaging from $4.79$ to $5.57$ turns, NQP-Bench preserves interaction complexity, presenting a challenge for memory-driven reasoning.

\begin{table}[t]
\centering
\small
\renewcommand{\arraystretch}{1.08}
\begin{tabular*}{0.82\linewidth}{@{\extracolsep{\fill}}lcccc}
\toprule
Subset & Train & Test & Avg. Turns & Sources \\
\midrule
NQP-Priv & 21{,}707 & 2{,}212 & 4.79 & Private \\
NQP-Wild & 15{,}916 & 2{,}354 & 4.82 & WildChat \\
NQP-Share & -- & 1{,}947 & 5.57 & ShareChat \\
\bottomrule
\end{tabular*}
\caption{
Statistics of the NQP-Bench datasets. The benchmark provides a robust mix of private and public data for training, while reserving a dedicated cross-source subset exclusively for testing generalization.
}
\label{tab:dataset_stats}
\end{table}

\paragraph{Intent Evaluation Rubric.}
Evaluating generative predictions in open-ended conversations is challenging. Traditional string-matching metrics penalize intent-equivalent queries with different surface forms, while binary judgments fail to capture partial predictive success, such as predicting the correct topic but diverging in the specific question. To quantify partial contextual understanding, we introduce a 5-point Likert scale~\citep{likert1932technique}, shown in Table~\ref{tab:judge_rubric}, based on our scoring protocol. This rubric accommodates the diversity of valid user intents and serves as the primary evaluation metric for both human and automated evaluators.

\begin{table}[htbp]
\centering
\small
\setlength{\tabcolsep}{3pt}
\renewcommand{\arraystretch}{1.0}
\begin{tabular*}{0.82\linewidth}{@{}c@{\hspace{2pt}}l@{}}
\toprule
Score & Criterion \\
\midrule
5 & \textbf{Intent hit}: same core intent, equivalent phrasing \\
4 & \textbf{Highly aligned}: same intent, different phrasing \\
3 & \textbf{Topic related}: shared topic, different specific question \\
2 & \textbf{Slightly related}: same domain, different direction \\
1 & \textbf{Irrelevant}: no meaningful connection \\
\bottomrule
\end{tabular*}
\caption{Intent evaluation rubric for scoring predicted queries against the ground truth.}
\label{tab:judge_rubric}
\end{table}

\begin{figure*}[t]
    \centering
    \includegraphics[width=0.80\textwidth]{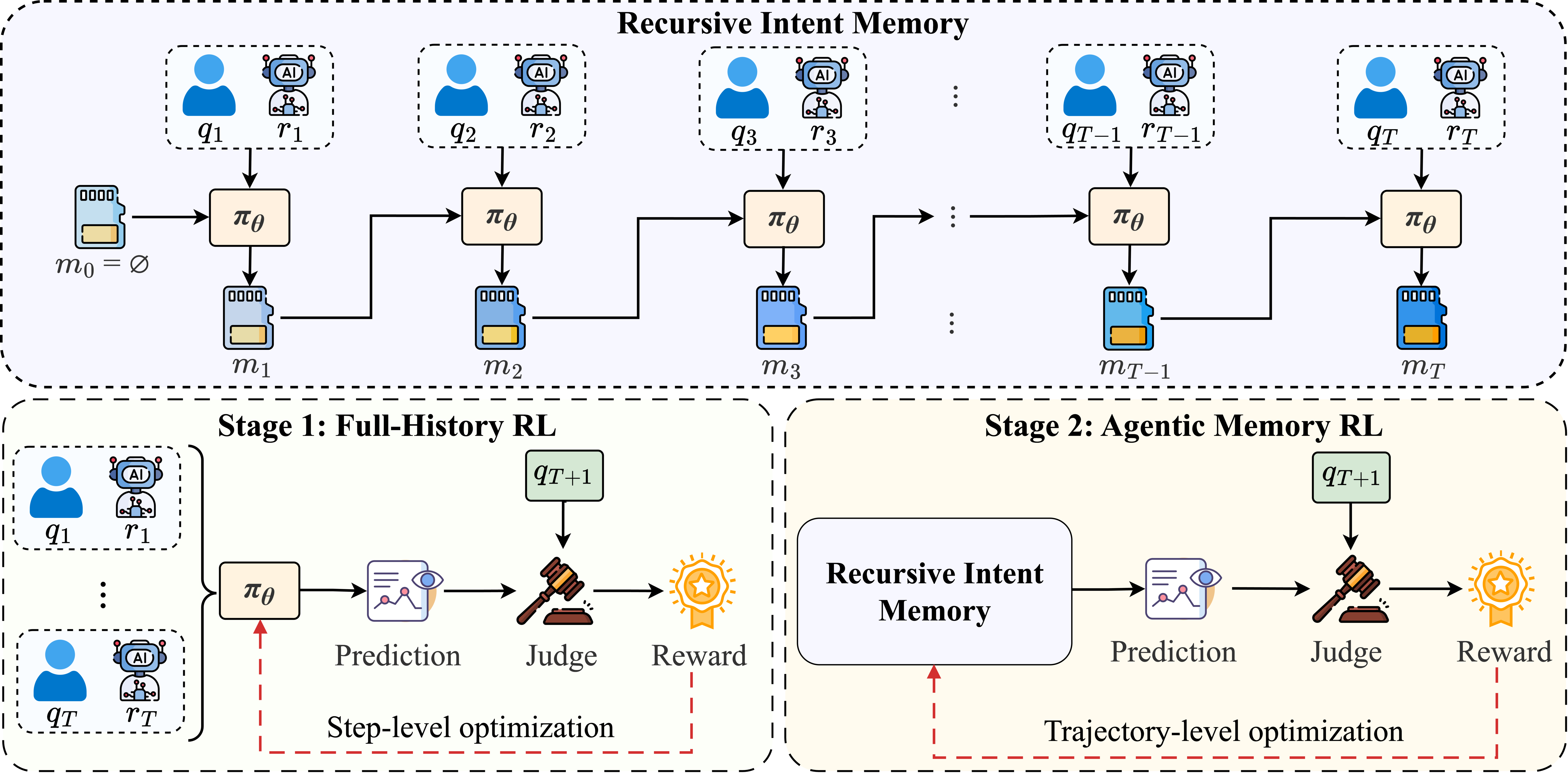} 
    \caption{Overview of OnePred. \textbf{Top}: the recursive intent memory mechanism. At each turn, the model receives only the previous memory $m_{t-1}$ and the current observation $(q_t, r_t)$, and outputs an updated memory $m_t$. \textbf{Bottom}: the two-stage RL training pipeline. Stage~1 (Full-History RL) treats the entire conversation as a single-step input and directly optimizes prediction. Stage~2 (Agentic Memory RL) trains the model to predict through its memory alone over a multi-turn trajectory, broadcasting the final-turn reward to all preceding memory-update steps.}
    \label{fig:method_overview}
\end{figure*}

\section{Method}
\label{sec:method}

OnePred consists of two components: a \emph{recursive intent memory}, which is incrementally updated at each turn to maintain prediction-relevant context (\S\ref{sec:memory}), and a \emph{two-stage RL training pipeline} designed to optimize both the model's predictive ability and its memory compression strategy (\S\ref{sec:training}).
Figure~\ref{fig:method_overview} provides an overview.

\subsection{Recursive Intent Memory}
\label{sec:memory}

OnePred maintains a text-based memory state $m_t$ that undergoes incremental updates as the conversation unfolds. At each turn, the model is conditioned exclusively on the previous memory state and the current user--assistant exchange, and decides which information to retain, update, or discard.
Crucially, the raw conversation history is not accessible; the memory serves as the sole conduit for propagating historical context to the final prediction step.
The memory is further bounded by a maximum length of $k$ tokens, with any surplus content truncated before the next turn.
This combination of memory-only context and token cap creates a hard information bottleneck, forcing the model to distill and retain only context that is useful for prediction.

Unlike a general-purpose summary that compresses information without a task-specific objective, OnePred's memory is shaped by the prediction reward (\S\ref{sec:reward}) to selectively retain prediction-relevant signals, forming what we term an \textbf{intent chain}. The intent chain is represented as free-form text: its content is not prescribed by a fixed schema, but emerges through RL optimization to encode the user's evolving topics and active needs.

Formally, given a $T$-turn conversation with observations $o_t = (q_t, r_t)$, the model takes the previous memory and the current observation as input, and generates an updated memory together with $N$ candidate predictions:
\begin{equation}
m_t,\; \hat{q}_t^{(1\mkern-2mu)}\!, \ldots, \hat{q}_t^{(N\mkern-2mu)} = f_\theta\!\bigl(m_{t-1}, o_t\bigr),
\label{eq:inference}
\end{equation}
where $t = 1, \ldots, T$, the initial memory $m_0$ is empty, and $f_\theta$ denotes the language model, with no additional parametric module.
Since the memory is text-based, the same LLM can both read and write it, while keeping its content directly interpretable (see Appendix~\ref{app:case_study}). The model outputs $N$ candidates rather than a single prediction to account for the open-ended nature of multi-turn dialogue, where multiple plausible continuations may exist (see \S\ref{sec:reward} for scoring).

This mechanism is used differently during inference and training.
At inference time, the model emits both a memory update and candidate predictions after every turn, bounding the input size by at most $k + |o_t|$ tokens regardless of conversation length. This contrasts with Full-history inputs, whose context grows linearly with $T$.
During training, the model updates memory at each intermediate turn ($t = 1, \ldots, T{-}1$) and emits candidate predictions only at the final turn $T$. Intermediate memory updates receive no direct supervision, but are shaped indirectly through the final-turn reward (\S\ref{sec:reward}).

\subsection{Two-Stage Training}
\label{sec:training}

The training objective is to optimize a single model $f_\theta$ that simultaneously maintains effective memory and produces accurate predictions.
Supervised fine-tuning is not well suited to this task for two reasons. First, there is no ground-truth annotation specifying what the memory should contain. Second, next-query prediction is inherently open-ended, so requiring an exact match to a single reference would unfairly penalize semantically equivalent outputs. We therefore use reinforcement learning with an LLM-as-a-judge reward that assesses intent equivalence, allowing the final prediction reward to shape the memory content generated at earlier turns.

End-to-end memory training requires the model to learn both \emph{what} is predictive and \emph{how} to compress it into bounded memory. These two abilities are interdependent: effective compression requires knowing which signals matter for prediction, while learning to predict through memory requires the memory to already contain useful information. To break this circular dependency, we decompose training into two sequential stages, both optimized with Group Relative Policy Optimization (GRPO;~\citealt{shao2024deepseekmath}).

\paragraph{Stage~1: Learning to Predict (Full-History RL).}
We first train the model to predict $q_{T+1}$ from the complete conversation history, bypassing the memory mechanism entirely. This stage allows the model to identify predictive signals, such as topic shifts and unresolved questions, without the additional difficulty of memory management. It thereby provides a strong initialization for the subsequent memory-learning stage.

\paragraph{Stage~2: Learning to Compress (Agentic Memory RL).}
Starting from the Stage~1 checkpoint, we switch to the memory interface defined in Eq.~\ref{eq:inference}. The model now processes turns sequentially, conditioned exclusively on its previously generated memory and the current observation.
Each rollout constitutes a $T$-step trajectory: for turns $1$ through $T-1$, the model updates the memory state; at turn $T$, it outputs the final memory update together with $N$ candidate predictions.
The prediction objective and reward remain identical to Stage~1, but the full history is no longer accessible.
The model is therefore forced to learn to preserve its predictive knowledge through bounded memory.

The two stages are complementary: Stage~1 establishes what to predict, while Stage~2 teaches how to compress the necessary context into bounded memory.
Our ablation study (\S\ref{sec:ablation}) confirms that neither stage alone matches the full pipeline.

\paragraph{Reward Design.}
\label{sec:reward}
Given $N$ candidate predictions, the reward evaluates how well they capture the user's actual next query $q_{T+1}$.
We adopt a best-of-$N$ strategy: each candidate $\hat{q}^{(i)}$ is independently scored against the ground truth, and the highest score is taken as $R_{\text{judge}}$.
The reward combines two components:
\begin{equation}
R = \lambda \cdot R_{\text{judge}} + (1 - \lambda) \cdot R_{\text{format}}.
\label{eq:reward}
\end{equation}
$R_{\text{judge}}$ is computed by an ensemble of LLM judges. Each judge independently rates the candidate--ground-truth pair on a discrete intent-alignment scale, and the final score is determined by majority vote (details in \S\ref{sec:setup}). This avoids relying on surface-level text-overlap metrics, such as BLEU and ROUGE, which cannot reliably capture intent equivalence between differently worded queries. $R_{\text{format}}$ penalizes structural violations, such as missing delimiters or malformed outputs, ensuring that the agent loop remains well formed throughout training. The coefficient $\lambda$ controls the trade-off between prediction quality and format compliance.

\paragraph{Credit Assignment.}
Both stages are optimized with the GRPO objective.
For a given prompt $x$, we sample a group of $G$ rollouts $\{y_1, \ldots, y_G\}$ from the sampling policy $\pi_{\theta_{\text{old}}}$.
The importance ratio between the updated policy $\pi_\theta$ and the sampling policy is defined as:
\begin{equation}
\rho_i = \frac{\pi_\theta(y_i \mid x)}{\pi_{\theta_{\text{old}}}(y_i \mid x)},
\label{eq:ratio}
\end{equation}
and the policy is optimized by minimizing
\begin{equation}
\mathcal{L}(\theta) = -\frac{1}{G}\!\sum_{i=1}^{G} \min\!\bigl(\rho_i \hat{A}_i,\; \bar\rho_i\, \hat{A}_i\bigr),
\label{eq:grpo}
\end{equation}
where $\bar\rho_i = \mathrm{clip}(\rho_i,\, 1{-}\epsilon,\, 1{+}\epsilon)$.
In Stage~1, $x$ is the full conversation history; in Stage~2, $x$ is the concatenation of the memory and the current observation at each turn. The advantage is normalized within each group:
\begin{equation}
\hat{A}_i = \frac{R_i - \operatorname{mean}(\{R_j\}_{j=1}^{G})}{\operatorname{std}(\{R_j\}_{j=1}^{G})}.
\label{eq:advantage}
\end{equation}

In Stage~1, each rollout consists of a single prediction step, so $\hat{A}_i$ is directly applied.
In Stage~2, each rollout is a multi-turn trajectory. Since the reward is only observed at the final prediction step $T$, yet heavily depends on the quality of earlier memory updates, we broadcast the trajectory-level advantage $\hat{A}_T$ uniformly to all preceding turns:
\begin{equation}
\hat{A}_t = \hat{A}_T, \quad t = 1, \ldots, T{-}1,
\label{eq:credit}
\end{equation}
so that every memory-update step receives gradient signal from the eventual prediction outcome.
This uniform broadcast is justified by the sequential dependency among memory states: each $m_t$ causally influences all subsequent memories and the final prediction, making earlier updates jointly responsible for the final prediction quality.

\section{Experiments}
\label{sec:experiments}

We evaluate our method along five dimensions: overall prediction quality (\S\ref{sec:main_results}), the contribution of each training stage (\S\ref{sec:ablation}), model scaling (\S\ref{sec:scaling}), inference efficiency (\S\ref{sec:efficiency}), and robustness to dialogue length (\S\ref{sec:turn_analysis}). A qualitative case study is provided in Appendix~\ref{app:case_study}.

\subsection{Experimental Setup}
\label{sec:setup}

We evaluate on the three NQP-Bench subsets (\S\ref{sec:benchmark}) and compare three history interfaces: \textbf{Current-turn}, which uses only the most recent exchange; \textbf{Full-history}, which concatenates all preceding turns; and \textbf{Ours}, which uses recursive memory. Each interface is evaluated under three model regimes: \textbf{Gemini-3.1-Pro}~\citep{gemini31pro_modelcard}, a frontier closed-source model; \textbf{Base Qwen}, Qwen3-8B~\citep{qwen3} without task-specific training; and \textbf{RL-trained Qwen}, Qwen3-8B optimized with two-stage RL.
For the RL-trained regime, we train a separate checkpoint for each history interface on the NQP-Priv and NQP-Wild training splits. Specifically, the Current-turn and Full-history models are optimized using Stage~1 RL on their respective inputs, while our memory model follows the complete two-stage pipeline described in \S\ref{sec:training}. NQP-Share is held out entirely during training, serving exclusively as an out-of-distribution (OOD) generalization test.
The primary metric is the \textbf{LLM Judge} score, based on the rubric in Table~\ref{tab:judge_rubric} and mapped to $[0,100]$. The final score is determined by majority vote among three commercial judges. We also report \textbf{Human} scores from expert annotators. Training details are provided in Appendix~\ref{sec:training_details}.

\subsection{Main Results}
\label{sec:main_results}

\begin{table*}[t]
\centering
\small
\renewcommand{\arraystretch}{1.08}
\begin{tabular*}{0.92\textwidth}{@{\extracolsep{\fill}}>{\raggedright\arraybackslash}m{3.0cm}cccccccc}
\toprule
\multirow{2}{3.0cm}{\centering \textbf{Method}}
& \multicolumn{2}{c}{\textbf{NQP-Priv}}
& \multicolumn{2}{c}{\textbf{NQP-Wild}}
& \multicolumn{2}{c}{\textbf{NQP-Share}}
& \multicolumn{2}{c}{\textbf{Average}} \\
\cmidrule(lr){2-3} \cmidrule(lr){4-5} \cmidrule(lr){6-7} \cmidrule(lr){8-9}
& Judge & Human
& Judge & Human
& Judge & Human
& Judge & Human \\
\midrule

\multicolumn{9}{l}{\textbf{Gemini-3.1-Pro}} \\
\hspace{1em}Current-turn  &41.18  & 42.10 & 45.53 & 45.31 & 42.78 & 43.33 & \cellcolor{gray!15}43.16 & \cellcolor{gray!15}43.58 \\
\hspace{1em}Full-history  &40.81  & 41.15 & 48.81 & 49.86 & 46.30 & 46.13 & \cellcolor{gray!15}45.31 & \cellcolor{gray!15}45.71 \\
\hspace{1em}Ours          &\textbf{44.22}  & \textbf{44.85} & \textbf{50.72} & \textbf{51.14} & \textbf{48.77} & \textbf{49.66} & \cellcolor{gray!15}\textbf{47.90} & \cellcolor{gray!15}\textbf{48.55} \\
\midrule

\multicolumn{9}{l}{\textbf{Base Qwen}} \\
\hspace{1em}Current-turn  &28.92  & 30.07 & 36.09 & 36.42 & 35.02 & 35.86 & \cellcolor{gray!15}33.34 & \cellcolor{gray!15}34.12 \\
\hspace{1em}Full-history  &31.37  & 31.85 & 38.20 & 39.16 & 35.69 & 36.30 & \cellcolor{gray!15}35.09 & \cellcolor{gray!15}35.77 \\
\hspace{1em}Ours          &\textbf{33.46}  & \textbf{33.18} & \textbf{39.57} & \textbf{40.42} & \textbf{37.84} & \textbf{37.62} & \cellcolor{gray!15}\textbf{36.96} & \cellcolor{gray!15}\textbf{37.07} \\
\midrule

\multicolumn{9}{l}{\textbf{RL-trained Qwen}} \\
\hspace{1em}Current-turn  &38.17  & 38.63 & 44.02 & 44.20 & 40.50 & 40.22 & \cellcolor{gray!15}40.90 & \cellcolor{gray!15}41.02 \\
\hspace{1em}Full-history  &40.79  & 40.64 & 44.35 & 44.97 & 42.16 & 42.53 & \cellcolor{gray!15}42.43 & \cellcolor{gray!15}42.71 \\
\hspace{1em}Ours          &\textbf{42.15}  & \textbf{42.72} & \textbf{46.00} & \textbf{47.27} & \textbf{43.84} & \textbf{44.96} & \cellcolor{gray!15}\textbf{44.00} & \cellcolor{gray!15}\textbf{44.98} \\
\bottomrule
\end{tabular*}
\caption{
Main results on the three NQP-Bench subsets.
For each subset, we report both the LLM-judge score (Judge) and the human evaluation score (Human).
Rows are grouped by model regime: Gemini-3.1-Pro, Base Qwen, and RL-trained Qwen.
The last two columns report metric-wise averages across the three subsets.
}
\label{tab:main_results}
\end{table*}

Table~\ref{tab:main_results} presents the main comparison across three model regimes and three NQP-Bench subsets.

\paragraph{Consistent advantage across model regimes.}
Despite using only a bounded memory ($\leq\!k$ tokens) instead of the full conversation history, OnePred outperforms Full-history in all nine cells (three regimes $\times$ three datasets).
Averaged across datasets, the Judge-score margin over Full-history is $+2.6$ for Gemini-3.1-Pro, $+1.9$ for Base Qwen, and $+1.6$ for RL-trained Qwen.
The gap is largest for Gemini-3.1-Pro and smallest for RL-trained Qwen, which is expected because RL training benefits all history interfaces. For example, Full-history improves from $35.09$ with Base Qwen to $42.43$ with RL-trained Qwen, narrowing the absolute gap while preserving the overall ranking.
The consistent gains across model capability levels, from an untrained 8B model to a frontier commercial API, suggest that the advantage stems from the memory-based history interface rather than model-specific artifacts.
Human evaluation yields the same ranking in all nine comparisons, supporting the reliability of the LLM-judge protocol.
We further compare against sliding-window and summarize-then-predict baselines in Appendix~\ref{app:additional_baselines}, conduct hyperparameter sensitivity analyses in Appendix~\ref{app:hyperparameter}, report bootstrap significance tests in Appendix~\ref{app:significance}, analyze fine-grained performance in Appendix~\ref{app:failure}, and provide rubric-level score distributions in Appendix~\ref{app:score_dist}.

\paragraph{Private benchmark (NQP-Priv).}
On NQP-Priv, which reflects realistic deployment conditions with genuine user conversations, OnePred achieves the highest score in every model regime.
The largest single margin appears under Gemini-3.1-Pro ($+3.4$ over Full-history), showing that recursive memory effectively captures cross-turn intent patterns in real user behavior.
Notably, under Gemini-3.1-Pro, Current-turn and Full-history perform comparably ($41.18$ vs.\ $40.81$), suggesting that naively concatenating noisy real-world dialogue history does not reliably improve prediction and may even slightly degrade performance. This observation further motivates a selective memory mechanism that retains predictive signals while filtering out noise.

\begin{table}[t]
\centering
\small
\renewcommand{\arraystretch}{1.08}
\begin{tabular*}{0.65\linewidth}{@{\extracolsep{\fill}}>{\raggedright\arraybackslash}m{3.6cm}cc}
\toprule
Method & Judge & Human \\
\midrule
Base Qwen (no training)         &39.57  & 40.42 \\
+ Stage 1 only      &43.82 & 44.63 \\
+ Stage 2 only      &42.96  & 43.35 \\
+ Stage 1 then Stage 2    &46.00 & 47.27 \\
\bottomrule
\end{tabular*}
\caption{Ablation of the two-stage training pipeline on NQP-Wild. All rows are evaluated with the recursive memory interface; only the training differs. Stage~1 only trains with full-history input but is evaluated under memory prompting without memory-specific RL.}
\label{tab:ablation}
\end{table}

\paragraph{Public and cross-source benchmarks (NQP-Wild, NQP-Share).}
The same pattern holds on both public benchmarks.
On NQP-Wild, our method leads Full-history by $+1.9$ (Gemini), $+1.4$ (Base), and $+1.7$ (RL-trained).
Full-history benefits more from the longer, structured conversations in NQP-Wild than from noisy private logs ($48.81$ vs.\ $40.81$ under Gemini), yet our method still outperforms it.
On NQP-Share, which tests cross-source generalization, the advantage persists ($+2.5$ under Gemini and $+1.7$ under RL-trained). These results indicate that the learned memory interface transfers beyond the training distribution.

\begin{table}[t]
\centering
\small
\renewcommand{\arraystretch}{1.08}
\begin{tabular*}{0.76\linewidth}{@{\extracolsep{\fill}}lcccc}
\toprule
\multirow{2}{*}{Size}
& \multicolumn{2}{c}{Ours (Base)}
& \multicolumn{2}{c}{Ours (RL-trained)} \\
\cmidrule(lr){2-3} \cmidrule(lr){4-5}
& Judge & Human & Judge & Human \\
\midrule
1.7B & 29.24 & 30.47 & 36.54 & 36.94 \\
4B & 34.07 & 33.88 & 42.34 & 42.09 \\
8B & 39.57 & 40.42 & 46.00 & 47.27  \\
\bottomrule
\end{tabular*}
\caption{Scaling results on NQP-Wild with 1.7B, 4B, and 8B Qwen3 backbones.}
\label{tab:scaling}
\end{table}

\subsection{Ablation on the Two-Stage Training Pipeline}
\label{sec:ablation}

Table~\ref{tab:ablation} ablates the contribution of each training stage on NQP-Wild.

\paragraph{Stage~1: learning to predict.}
Full-History RL alone brings a substantial improvement over the untrained Base Qwen ($39.57 \to 43.82$, $+4.3$).
By training with complete conversation context, Stage~1 directly improves the model's next-query prediction ability and provides a strong initialization for subsequent memory training.

\paragraph{Stage~2: learning to compress.}
Agentic Memory RL from the Base model also yields meaningful gains ($39.57 \to 42.96$, $+3.4$), confirming that learning to predict through a compact memory is beneficial even without full-history pre-training.

\paragraph{Two-stage: complementary gains.}
The full two-stage pipeline achieves the best judge score of $46.00$, exceeding Stage~1 alone by $+2.2$ and Stage~2 alone by $+3.0$.
The gains are complementary: prediction skill from Stage~1 and compression skill from Stage~2 address different bottlenecks.

\subsection{Scaling Analysis}
\label{sec:scaling}

Table~\ref{tab:scaling} examines how OnePred scales with model size on NQP-Wild, using Qwen3 backbones with 1.7B, 4B, and 8B parameters.
Without task-specific training, performance increases monotonically with scale: $29.24$ (1.7B) $\to$ $34.07$ (4B) $\to$ $39.57$ (8B), indicating that larger capacity helps maintain coherent memory and predict intent.
The two-stage RL pipeline yields consistent gains across all sizes: $+7.3$ (1.7B), $+8.3$ (4B), and $+6.4$ (8B). The largest absolute gain appears at 4B, while the 8B model already captures more predictive patterns from pretraining alone.

\begin{figure}[t]
\centering
\includegraphics[width=0.76\linewidth]{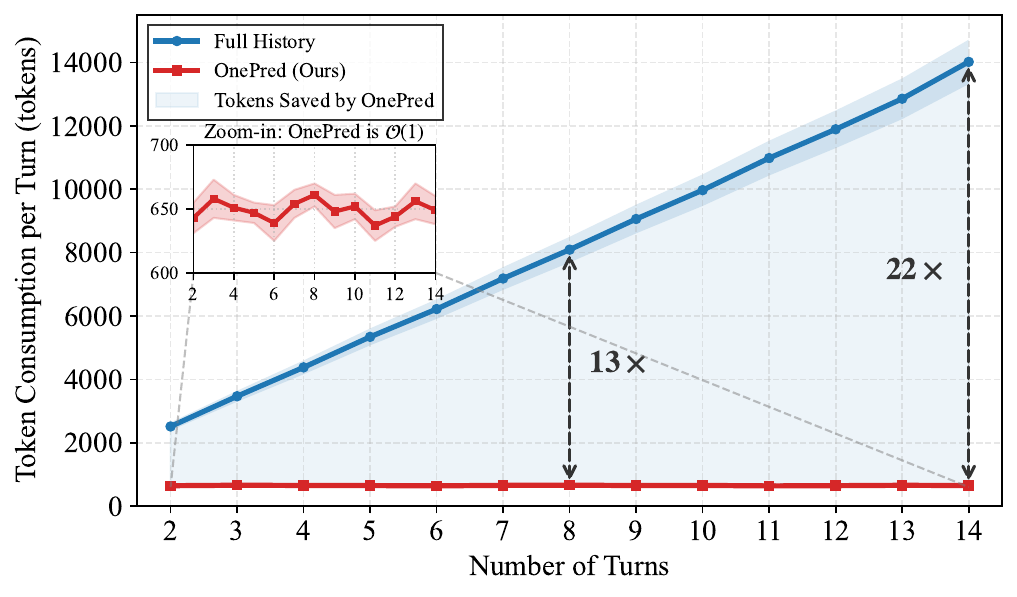}
\caption{Average input tokens per turn for Full-history and OnePred across dialogue lengths on NQP-Wild. Full-history grows linearly with dialogue length, whereas OnePred remains bounded.}
\label{fig:efficiency}
\end{figure}

\subsection{Inference Efficiency}
\label{sec:efficiency}

OnePred substantially reduces inference cost. In a deployed system, next-query prediction runs after each turn, since the predictor must generate candidate queries before the user types.
Full-history and OnePred therefore require the same number of LLM calls (one per turn); the difference lies in the input size of each call.
For Full-history, the prompt at turn $t$ includes all $t$ previous exchanges, causing the context length to grow linearly with conversation length. In contrast, OnePred conditions on the system prompt, bounded memory ($\leq\!k$ tokens), and the current observation, making the input size independent of the number of prior turns.

Figure~\ref{fig:efficiency} quantifies this difference on NQP-Wild.
Our method uses roughly $650$ tokens per turn regardless of conversation length.
Full-history starts at ${\sim}2{,}500$ tokens for a 2-turn dialogue and grows to over $14{,}000$ tokens by turn~14, reaching a $13\times$ gap at turn~8 and $22\times$ at turn~14.
This gap persists even with KV caching, since each generated token still attends to all cached states, keeping decode-time cost proportional to sequence length. By bounding the input, OnePred ensures consistent decode latency regardless of conversation length.

\subsection{Performance by Dialogue Length}
\label{sec:turn_analysis}

\begin{figure}[t]
\centering
\includegraphics[width=0.76\linewidth]{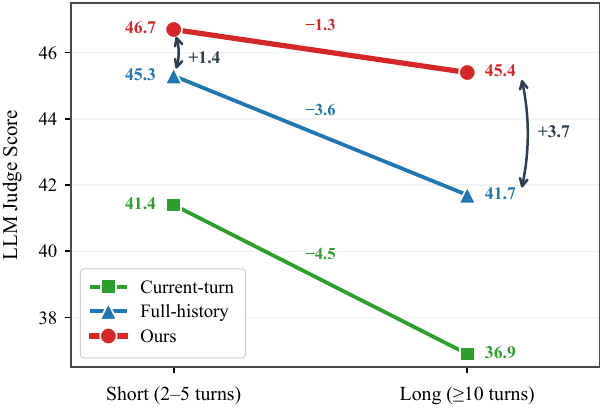}
\caption{Performance by dialogue length on NQP-Wild with RL-trained Qwen. OnePred retains $97\%$ performance on long dialogues ($\geq$10 turns), while Full-history and Current-turn degrade more sharply.}
\label{fig:turn_analysis}
\end{figure}

To analyze how each method handles longer conversations, we split NQP-Wild into short conversations with 2--5 turns and long conversations with at least 10 turns, and report the Judge score of RL-trained Qwen for each group in Figure~\ref{fig:turn_analysis}.

All methods perform worse on long conversations, but the degree of degradation differs substantially. Current-turn drops by $4.5$ points ($41.4 \to 36.9$), as a single exchange provides increasingly insufficient context. Full-history drops by $3.6$ points ($45.3 \to 41.7$), likely because longer prompts introduce more irrelevant context and dilute predictive signals. In contrast, OnePred drops by only $1.3$ points ($46.7 \to 45.4$), retaining $97\%$ of its short-conversation performance. Its advantage over Full-history widens from $+1.4$ points on short conversations to $+3.7$ points on long conversations, indicating that recursive memory is particularly effective as conversations grow longer.

\section{Conclusion}
We formalize next-query prediction in multi-turn LLM conversations and introduce OnePred, which maintains a recursively updated intent memory for proactive prediction without full-history concatenation. Experiments on NQP-Bench show that OnePred reduces per-turn input tokens by up to $22\times$ compared with Full-history while outperforming all baselines in prediction quality. The bounded memory acts as a task-oriented information bottleneck that filters conversational noise, yielding robust performance as conversations lengthen. We release the public subsets of NQP-Bench and OnePred as a scalable, interpretable step from reactive response generation toward proactive interaction.

\section*{Limitations}

The bounded memory keeps per-turn cost constant but inevitably loses some fine-grained details during compression, such as specific numbers, exact phrasing, or minor sub-topics mentioned in passing. Appendix~\ref{app:failure} shows cases where such details become relevant to the next query. Our RL training and open-source experiments use Qwen3 backbones ranging from 1.7B to 8B. Although we also evaluate Gemini-3.1-Pro, we have not tested other backbone families such as Llama or Mistral. Finally, LLMs are involved in both benchmark curation and evaluation scoring. Although we use architecturally distinct models for these two stages and validate curation quality with human annotators ($\kappa = 0.83$; Appendix~\ref{app:judge_validation}), some residual shared bias between LLM curation and LLM evaluation cannot be fully ruled out. Human evaluation yields the same method ranking across all comparisons, but the two protocols may still disagree on individual samples.

\section*{Ethical Considerations}
Our work uses two public datasets under their respective terms: WildChat~\citep{zhao2024wildchat}, which was collected with users' informed consent and released after PII removal, and ShareChat~\citep{sharechat-2025}, which was collected from user-shared conversation URLs under IRB approval and de-identified using Microsoft Presidio with auxiliary model verification. Our curation pipeline only selects predictable conversation prefixes without altering the original content. The private NQP-Priv subset is derived from internal deployment logs and is not released; all results are reported in aggregate, with no individual conversations or user identifiers exposed. The use of these logs complies with the platform's terms of service, under which users consented to anonymized data being used for service improvement and research. All conversations were de-identified prior to annotation, and the study was reviewed and approved under the organization's internal data governance policy. Human annotators were informed of the task purpose and compensated at local market rates. Annotators were recruited from graduate students or professional annotators with experience in dialogue evaluation, and no personally identifying information about annotators is released. Because next-query prediction models user intent trajectories and may be misused for profiling or manipulation, deployment should be limited to assistive settings with clear user consent, opt-out mechanisms, strict data governance, and safeguards against using predicted intents for advertising, behavioral profiling, or other non-assistive purposes.


\bibliographystyle{conference}
\bibliography{custom}

@inproceedings{zhao2024wildchat,
  title={WildChat: 1M ChatGPT Interaction Logs in the Wild},
  author={Zhao, Wenting and Ren, Xiang and Hessel, Jack and Cardie, Claire and Choi, Yejin and Deng, Yuntao},
  booktitle={Proceedings of the International Conference on Learning Representations (ICLR)},
  year={2024}
}

@article{sharechat-2025,
  title={ShareChat: A Dataset of Chatbot Conversations in the Wild},
  author={Yan, Yueru and Nguyen, Tuc and Su, Bo and Lieffers, Melissa and Le, Thai},
  journal={arXiv preprint arXiv:2512.17843},
  year={2025}
}

@inproceedings{zheng2024lmsys,
  title={LMSYS-Chat-1M: A Large-Scale Real-World LLM Conversation Dataset},
  author={Zheng, Lianmin and Chiang, Wei-Lin and Sheng, Ying and Zhuang, Siyuan and Wu, Zhanghao and Zhuang, Yonghao and Lin, Zi and Li, Zhuohan and Li, Dacheng and Xing, Eric P. and Zhang, Hao and Gonzalez, Joseph E. and Stoica, Ion},
  booktitle={Proceedings of the International Conference on Learning Representations (ICLR)},
  year={2024}
}

@inproceedings{dehghani2017learning,
  title={Learning to Attend, Copy, and Generate for Session-Based Query Suggestion},
  author={Dehghani, Mostafa and Rothe, Sascha and Alfonseca, Enrique and Fleury, Pascal},
  booktitle={Proceedings of the ACM International Conference on Information and Knowledge Management (CIKM)},
  year={2017}
}

@article{shao2024deepseekmath,
  title={DeepSeekMath: Pushing the Limits of Mathematical Reasoning in Open Language Models},
  author={Shao, Zhihong and Wang, Peiyi and Zhu, Qihao and Xu, Runxin and Song, Junxiao and Zhang, Mingchuan and Li, Y.K. and Wu, Y. and Guo, Daya},
  journal={arXiv preprint arXiv:2402.03300},
  year={2024}
}

@article{kang2026coreact,
  title={Co-ReAct: Rubrics as Step-Level Collaborators for ReAct Agents},
  author={Kang, Jiazheng and Zhang, Bowen and Song, Zixin and Chen, Jiangwang and Yang, Xiao and Zhu, Da and Jiang, Guanjun},
  journal={arXiv preprint arXiv:2605.23590},
  year={2026},
  eprint={2605.23590},
  archivePrefix={arXiv},
  primaryClass={cs.AI},
  url={https://arxiv.org/abs/2605.23590}
}

@article{liu2026mapd,
  title={From Proprietary to Open-Source: Bridging the Distribution Gap via Multi-Agent Protocol Distillation in Agentic Search},
  author={Liu, Junlin and Chen, Jiangwang and Song, Zixin and Zhou, Shuaiyu and Lv, Chunji and Wu, Hank and Jiang, Kailin and Wu, Jinyang and Yu, Bohan and Zhou, Chenxi},
  journal={arXiv preprint arXiv:2607.24280},
  year={2026},
  eprint={2607.24280},
  archivePrefix={arXiv},
  primaryClass={cs.AI},
  url={https://arxiv.org/abs/2607.24280}
}

@article{chen2026decoevo,
  title={DecoEvo: Score-Decoupled Co-Evolution of Solver and Rubric-Generator Skills in Text Space},
  author={Chen, Jiangwang and Song, Zixin and Liu, Junlin and Zhou, Shuaiyu and Wu, Haiyan and Shi, Haihan and Zhou, Chenxi and Li, Hanqing and Yang, Xiao and Zhu, Da and Jiang, Guanjun and Wan, Hai and Zhao, Xibin},
  journal={arXiv preprint arXiv:2607.25675},
  year={2026},
  eprint={2607.25675},
  archivePrefix={arXiv},
  primaryClass={cs.AI},
  url={https://arxiv.org/abs/2607.25675}
}

@article{sheng2024verl,
  title={VERL: An Extensible Framework for Post-Training of Large Language Models},
  author={Sheng, Guangming and Zhang, Chi and Ye, Zilingfeng and Wu, Xibin and Zhang, Wang and Qi, Penghui and Lin, Runqi and Guo, Junliang and Xu, Wei},
  journal={arXiv preprint arXiv:2409.19256},
  year={2024}
}

@article{qwen3,
  title={Qwen3 Technical Report},
  author={{Qwen Team}},
  journal={arXiv preprint arXiv:2505.09388},
  year={2025}
}

@inproceedings{baeza2004query,
  title={Query recommendation using query logs in search engines},
  author={Baeza-Yates, Ricardo and Hurtado, Carlos and Mendoza, Marcelo},
  booktitle={International conference on extending database technology},
  pages={588--596},
  year={2004},
  organization={Springer}
}

@inproceedings{cao2008context,
  title={Context-aware query suggestion by mining click-through and session data},
  author={Cao, Huanhuan and Jiang, Daxin and Pei, Jian and He, Qi and Liao, Zhen and Chen, Enhong and Li, Hang},
  booktitle={Proceedings of the 14th ACM SIGKDD international conference on Knowledge discovery and data mining},
  pages={875--883},
  year={2008}
}

@inproceedings{sordoni2015hierarchical,
  title={A hierarchical recurrent encoder-decoder for generative context-aware query suggestion},
  author={Sordoni, Alessandro and Bengio, Yoshua and Vahabi, Hossein and Lioma, Christina and Grue Simonsen, Jakob and Nie, Jian-Yun},
  booktitle={proceedings of the 24th ACM international on conference on information and knowledge management},
  pages={553--562},
  year={2015}
}

@inproceedings{min2025ctr,
  title={CTR-guided generative query suggestion in conversational search},
  author={Min, Erxue and Huang, Hsiu-Yuan and Yang, Xihong and Yang, Min and Jia, Xin and Wu, Yunfang and Cai, Hengyi and Wang, Junfeng and Wang, Shuaiqiang and Yin, Dawei},
  booktitle={Proceedings of the 2025 Conference on Empirical Methods in Natural Language Processing: Industry Track},
  pages={2624--2634},
  year={2025}
}

@inproceedings{yin2026clicks,
  title={From clicks to preference: A multi-stage alignment framework for generative query suggestion in conversational system},
  author={Yin, Junhao and Wang, Haolin and Bao, Peng and Xu, Ju and Wang, Yongliang},
  booktitle={Proceedings of the 32nd ACM SIGKDD Conference on Knowledge Discovery and Data Mining V. 1},
  pages={2539--2550},
  year={2026}
}

@inproceedings{guo2026onesug,
  title={Onesug: The unified end-to-end generative framework for e-commerce query suggestion},
  author={Guo, Xian and Chen, Ben and Wang, Siyuan and Yang, Ying and Cheng, Mingyue and Lei, Chenyi and Ding, Yuqing and Li, Han},
  booktitle={Proceedings of the AAAI Conference on Artificial Intelligence},
  volume={40(17)},
  pages={14774--14782},
  year={2026}
}

@article{wang2025llm,
  title={LLM-Driven Preference Data Synthesis for Proactive Prediction of the Next User Utterance in Human-Machine Dialogue},
  author={Wang, Jinqiang and Ning, Huansheng and Ding, Jianguo and Zhu, Tao and Chen, Liming and Nugent, Chris},
  journal={arXiv preprint arXiv:2601.09713},
  year={2025}
}

@inproceedings{wu2019proactive,
  title={Proactive human-machine conversation with explicit conversation goal},
  author={Wu, Wenquan and Guo, Zhen and Zhou, Xiangyang and Wu, Hua and Zhang, Xiyuan and Lian, Rongzhong and Wang, Haifeng},
  booktitle={Proceedings of the 57th Annual Meeting of the Association for Computational Linguistics},
  pages={3794--3804},
  year={2019}
}

@article{deng2023survey,
  title={A survey on proactive dialogue systems: Problems, methods, and prospects},
  author={Deng, Yang and Lei, Wenqiang and Lam, Wai and Chua, Tat-Seng},
  journal={arXiv preprint arXiv:2305.02750},
  year={2023}
}

@article{liu2024lost,
  title={Lost in the middle: How language models use long contexts},
  author={Liu, Nelson F and Lin, Kevin and Hewitt, John and Paranjape, Ashwin and Bevilacqua, Michele and Petroni, Fabio and Liang, Percy},
  journal={Transactions of the association for computational linguistics},
  volume={12},
  pages={157--173},
  year={2024}
}

@article{wang2023longmem,
  title={Augmenting language models with long-term memory},
  author={Wang, Weizhi and Dong, Li and Cheng, Hao and Liu, Xiaodong and Yan, Xifeng and Gao, Jianfeng and Wei, Furu},
  journal={Advances in Neural Information Processing Systems},
  volume={36},
  pages={74530--74543},
  year={2023}
}

@article{yu2025memagent,
  title={MemAgent: Reshaping Long-Context LLM with Multi-Conv RL-Based Memory Agent},
  author={Yu, Hongli and Chen, Tinghong and Feng, Jiangtao and Chen, Jiangjie and Dai, Weinan and Yu, Qiying and Zhang, Ya-Qin and Ma, Wei-Ying and Liu, Jingjing and Wang, Mingxuan},
  journal={arXiv preprint arXiv:2507.02259},
  year={2025}
}

@article{zhou2025mem1,
  title={Mem1: Learning to synergize memory and reasoning for efficient long-horizon agents},
  author={Zhou, Zijian and Qu, Ao and Wu, Zhaoxuan and Kim, Sunghwan and Prakash, Alok and Rus, Daniela and Zhao, Jinhua and Low, Bryan Kian Hsiang and Liang, Paul Pu},
  journal={arXiv preprint arXiv:2506.15841},
  year={2025}
}

@article{guo2025deepseek,
  title={Deepseek-r1: Incentivizing reasoning capability in llms via reinforcement learning},
  author={Guo, Daya and Yang, Dejian and Zhang, Haowei and Song, Junxiao and Wang, Peiyi and Zhu, Qihao and Xu, Runxin and Zhang, Ruoyu and Ma, Shirong and Bi, Xiao and Shao, Zhihong},
  journal={arXiv preprint arXiv:2501.12948},
  year={2025}
}

@article{chen2026localsug,
  title={LocalSUG: Geography-Aware LLM for Query Suggestion in Local-Life Services},
  author={Chen, Jinwen and Gong, Shuai and Zhang, Shiwen and Zhang, Zheng and Zhao, Yachao and Wang, Lingxiang and Zhou, Haibo and Zhan, Yuan and Lin, Wei and Zhang, Hainan},
  journal={arXiv preprint arXiv:2603.04946},
  year={2026}
}

@article{likert1932technique,
  title={A technique for the measurement of attitudes.},
  author={Likert, Rensis},
  journal={Archives of psychology},
  year={1932}
}

@article{zeng2026glm,
  title={Glm-5: from vibe coding to agentic engineering},
  author={Zeng, Aohan and Lv, Xin and Hou, Zhenyu and Du, Zhengxiao and Zheng, Qinkai and Chen, Bin and Yin, Da and Ge, Chendi and Huang, Chenghua and Xie, Chengxing and Tang, Jie},
  journal={arXiv preprint arXiv:2602.15763},
  year={2026}
}

@article{deng2025proactive,
  title={Proactive conversational ai: A comprehensive survey of advancements and opportunities},
  author={Deng, Yang and Liao, Lizi and Lei, Wenqiang and Yang, Grace Hui and Lam, Wai and Chua, Tat-Seng},
  journal={ACM Transactions on Information Systems},
  volume={43},
  number={3},
  pages={1--45},
  year={2025},
  publisher={ACM New York, NY}
}

@misc{gemini31pro_modelcard,
      title={Gemini 3.1 Pro Model Card},
      author={{Google DeepMind}},
      year={2026},
      howpublished={\url{https://storage.googleapis.com/deepmind-media/Model-Cards/Gemini-3-1-Pro-Model-Card.pdf}},
      note={Accessed: 2026-05}
}

@misc{anthropic2026claude,
  title={Claude Opus 4.7 System Card},
  author={{Anthropic}},
  year={2026},
  month={apr},
  howpublished={\url{https://www.anthropic.com/system-cards}},
  note={Model system card for Claude Opus 4.7}
}

@misc{openai2026gpt55,
  title={GPT-5.5 System Card},
  author={{OpenAI}},
  year={2026},
  howpublished={\url{https://openai.com/index/gpt-5-5-system-card/}},
  note={Accessed: 2026-05}
}

@article{comanici2025gemini,
  title={Gemini 2.5: Pushing the frontier with advanced reasoning, multimodality, long context, and next generation agentic capabilities},
  author={Comanici, Gheorghe and Bieber, Eric and Schaekermann, Mike and Pasupat, Ice and Sachdeva, Noveen and Dhillon, Inderjit and Blistein, Marcel and Ram, Ori and Zhang, Dan and Rosen, Evan},
  journal={arXiv preprint arXiv:2507.06261},
  year={2025}
}

\appendix
\clearpage

\section{Implementation Details}
\label{sec:training_details}

\paragraph{Training setup.}
The backbone model is Qwen3-8B, trained with the verl framework~\citep{sheng2024verl} for GRPO.
Both stages use the AdamW optimizer with a PPO clip ratio $\epsilon = 0.2$ (Eq.~\ref{eq:grpo}).
\textbf{Stage~1} (Full-History RL): 500 steps, constant learning rate $1\!\times\!10^{-6}$, batch size 32, 8 rollouts per prompt, maximum prompt length 8{,}192, maximum response length 3{,}072.
\textbf{Stage~2} (Agentic Memory RL): initialized from the Stage~1 checkpoint, 200 steps, learning rate $5\!\times\!10^{-7}$ with cosine schedule and 2\% warmup, batch size 32, 8 rollouts per prompt, maximum response length 4{,}096.
The increased response length in Stage~2 accommodates the multi-turn trajectory format (memory update at each intermediate turn, with prediction appended only at the final turn).
The reward follows Eq.~\ref{eq:reward} with $\lambda = 0.9$; the memory budget is $k = 500$ tokens per turn; the number of candidate predictions is $N = 3$.
Sensitivity analyses for $k$, $N$, and $\lambda$ are provided in Appendix~\ref{app:hyperparameter}.
For the scaling experiments (Table~\ref{tab:scaling}), the 1.7B and 4B models are trained with the same hyperparameters as the 8B model.

\paragraph{Compute budget.}
Stage~1 and Stage~2 training were conducted on 8 NVIDIA A100 80GB GPUs, taking approximately 256 GPU-hours in total for the 8B model (including both stages and all rollout generation).

\paragraph{Inference.}
During evaluation, all models decode greedily (temperature $0$).
The memory budget is enforced by hard truncation at $k$ tokens after each memory update; if the model's output exceeds $k$ tokens in the memory field, the suffix beyond $k$ tokens is discarded before the next turn.

\paragraph{Evaluation details.}
The LLM Judge score is determined by majority vote of three frontier models: Claude Opus 4.7~\citep{anthropic2026claude}, GPT-5.5~\citep{openai2026gpt55}, and Gemini-3.1-Pro~\citep{gemini31pro_modelcard}.
Each judge independently rates every candidate--ground-truth pair on the 5-point rubric (Table~\ref{tab:judge_rubric}), and the majority rating is taken as the final score (median when no majority exists), linearly mapped to $[0, 100]$ via $(s - 1) / 4 \times 100$ where $s \in \{1,2,3,4,5\}$. The same ensemble is used for the training reward $R_{\text{judge}}$.
For human evaluation, five expert annotators independently scored 250 randomly sampled instances per subset on the same rubric, and their majority vote is taken as the final human score.
Further validation of the judge protocol is provided in Appendix~\ref{app:judge_validation}.

\section{Related Work}
\label{sec:related_work}

\subsection{Query Suggestion and Proactive Dialogue}
Query suggestion (QS) has evolved from traditional search environments~\citep{baeza2004query, cao2008context, sordoni2015hierarchical, dehghani2017learning} to conversational AI. Recent generative frameworks extend QS to proactive conversational settings, but are often designed for constrained domains, such as e-commerce~\citep{guo2026onesug}, local services~\citep{chen2026localsug}, and click-through-rate maximization~\citep{min2025ctr, yin2026clicks}. Importantly, these QS systems typically formulate the task as preference alignment over implicit feedback logs to induce user clicks. This differs from our next-query prediction task, which aims to model the natural trajectory of user intent.

Proactive dialogue systems typically steer conversations toward system-defined goals, such as task completion~\citep{deng2023survey,wu2019proactive}. Rather than anticipating the user's autonomous intent, they focus on selecting the next optimal system action. Co-ReAct~\citep{kang2026coreact}, for example, learns step-level rubrics that guide an agent's next search action, whereas our task is to predict the user's next query. Among existing approaches, \citet{wang2025llm} is most closely related to our work, as it anticipates user queries through offline-synthesized intent trees. However, their approach relies on heuristic structures and still requires full dialogue history at inference time, leaving the computational bottleneck unresolved. In contrast, OnePred tracks dynamic user intents through a bounded, recursively updated memory chain, avoiding both static structures and full-history concatenation.

\subsection{Memory-Augmented LLMs and Reinforcement Learning}
Managing long-range context in multi-turn dialogues commonly relies on full-history concatenation or retrieval-augmented generation. Although effective in preserving context, these paradigms suffer from increasing inference costs and signal dilution~\citep{liu2024lost, wang2023longmem}. Recent work introduces RL to maintain bounded memory for factual retrieval and task execution across extended contexts~\citep{yu2025memagent, zhou2025mem1}. In agentic search, MAPD distills structured multi-agent exploration protocols into a single policy~\citep{liu2026mapd}. While these methods optimize information retention or search behavior, OnePred learns a forward-looking memory that retains evolving intent trajectories for next-query prediction.

Training such a memory further introduces a credit-assignment challenge. Standard RL applications, including GRPO~\citep{shao2024deepseekmath, guo2025deepseek}, are typically applied to single-step generation. Related work has also optimized persistent text-space artifacts: DecoEvo co-evolves solver and rubric-generator skills with score-decoupled objectives~\citep{chen2026decoevo}. In contrast, OnePred optimizes a sequence of memory states without intermediate supervision, where the final prediction depends on all preceding memory updates. To address this, OnePred broadcasts the final-turn trajectory advantage across preceding memory-update steps, adapting GRPO to multi-turn conversational prediction.

\begin{figure*}[ht]
    \centering
    \begin{subfigure}[b]{0.40\textwidth}
        \centering
        \includegraphics[width=\textwidth]{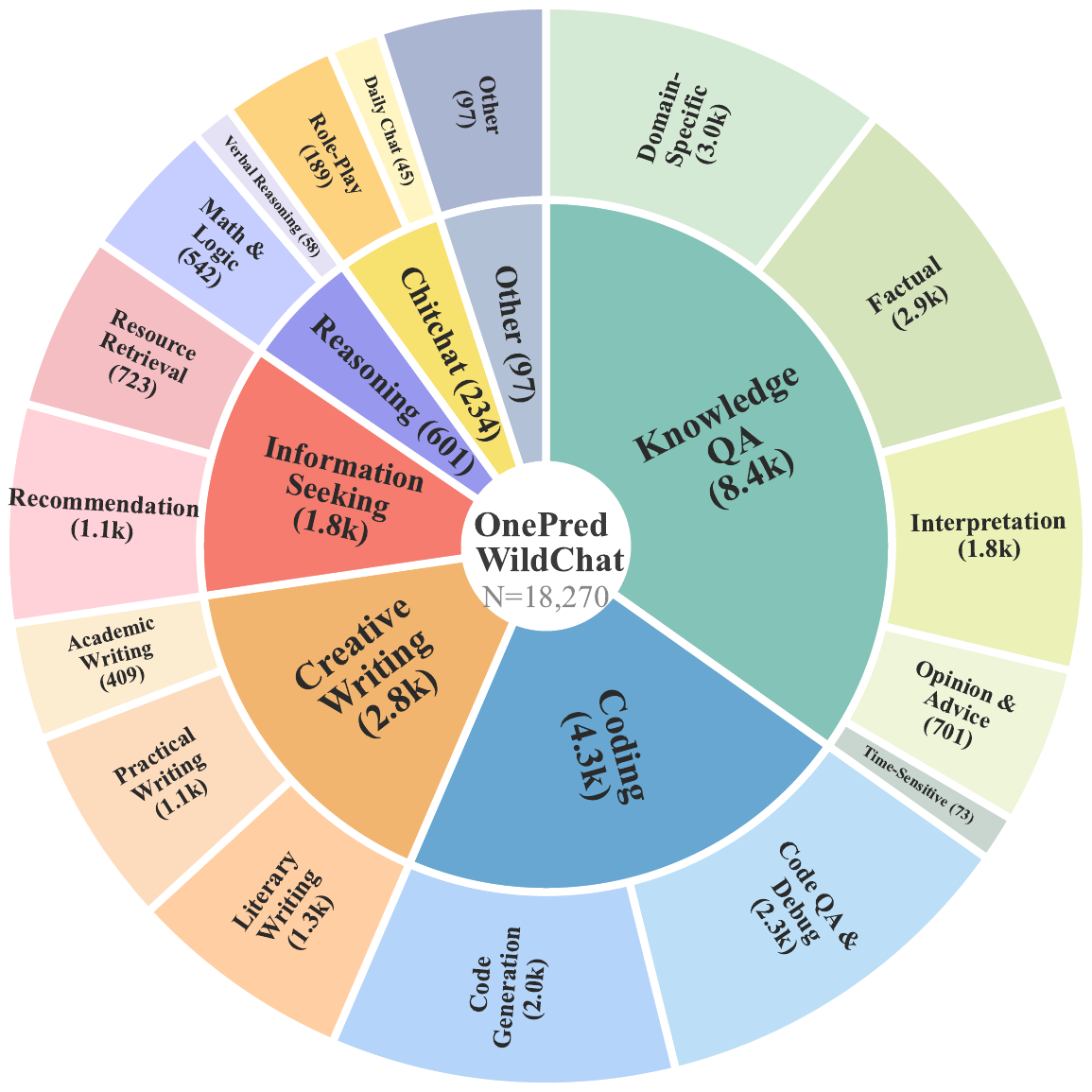}
        \caption{Intent Distribution (NQP-Wild)}
        \label{fig:intent_sunburst_wild}
    \end{subfigure}
    \hfill
    \begin{subfigure}[b]{0.40\textwidth}
        \centering
        \includegraphics[width=\textwidth]{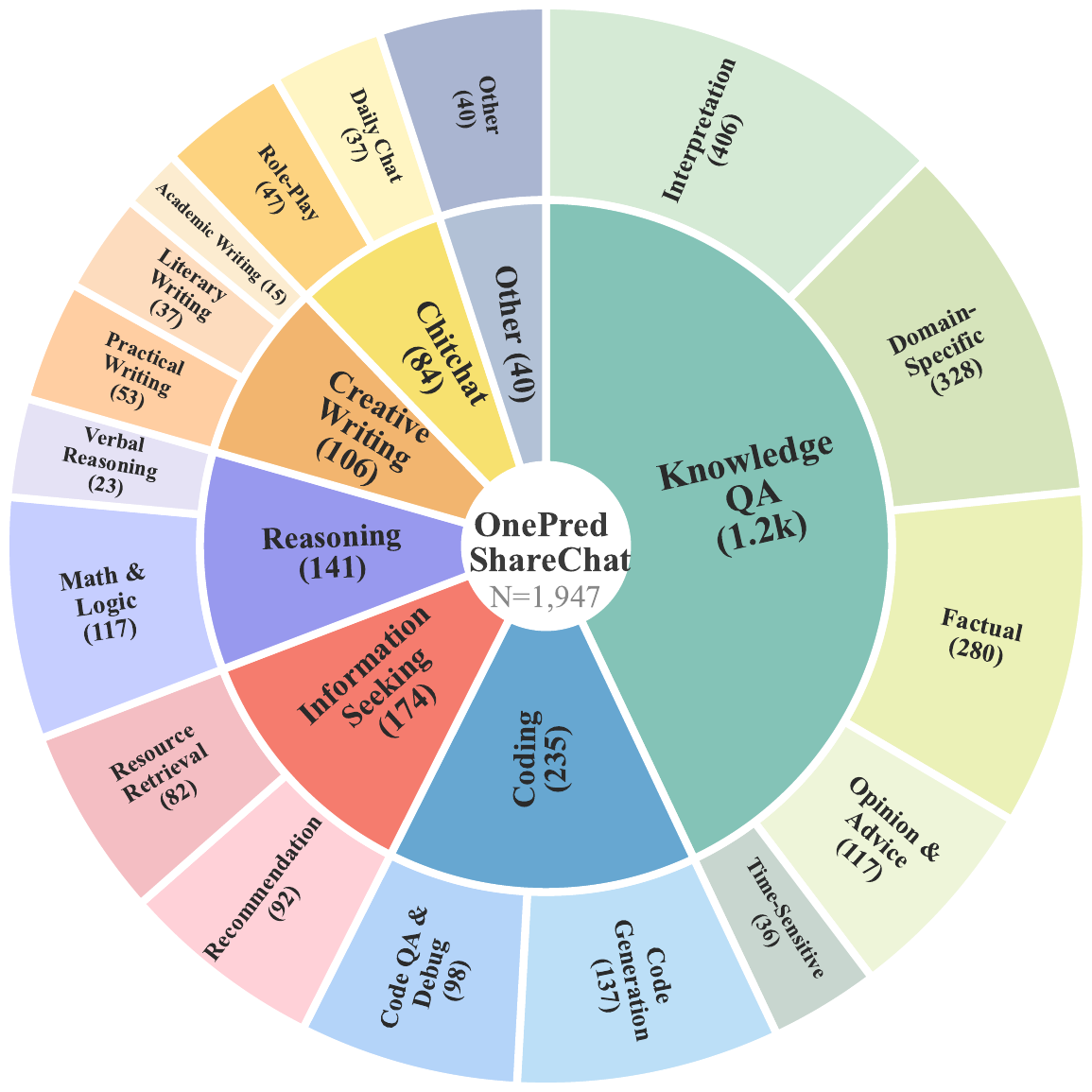}
        \caption{Intent Distribution (NQP-Share)}
        \label{fig:intent_sunburst_share}
    \end{subfigure}
    \caption{Sunburst charts detailing the hierarchical intent taxonomy and their distributions. The datasets demonstrate high diversity across knowledge-seeking, reasoning, and generative tasks.}
    \label{fig:intent_sunburst}
\end{figure*}

\section{Dataset Statistics and Characteristics}
\label{app:dataset_statistics}

This section presents the statistical characteristics of NQP-Bench. All structured metadata used in this analysis were automatically annotated using Gemini 3.1 Pro. Through visual analysis of intent distributions, dialogue lengths, intent transfer dynamics, and difficulty, we demonstrate that NQP-Bench is a highly diverse and challenging benchmark suitable for evaluating next-query prediction in multi-turn interactions.

\subsection{Intent Diversity}
NQP-Bench employs a fine-grained intent taxonomy consisting of 7 primary and 17 secondary intents. Figure~\ref{fig:intent_sunburst} illustrates the intent distribution across the WildChat and ShareChat subsets. The data spans a wide spectrum of user needs, anchored by three major domains: \textit{Knowledge QA} (focusing on factual queries and domain-specific interpretation), \textit{Coding} (spanning code generation and debugging), and \textit{Creative Writing}. This comprehensive distribution ensures that models are evaluated on their predictive capabilities across diverse real-world application scenarios, rather than overfitting to a single domain.

\begin{figure}[htbp]
    \centering
    \begin{subfigure}[b]{0.40\textwidth}
        \centering
        \includegraphics[width=\textwidth]{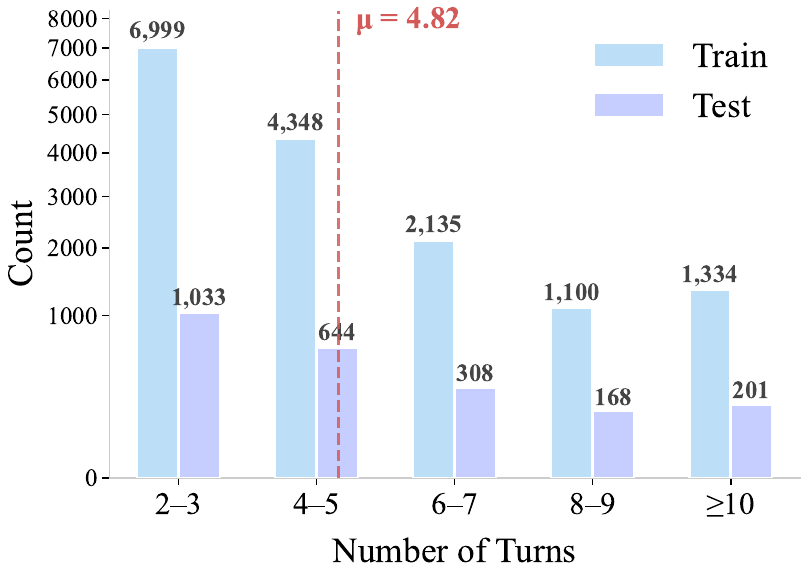}
        \caption{Turn Distribution (NQP-Wild)}
        \label{fig:turns_wild}
    \end{subfigure}
    \hfill
    \begin{subfigure}[b]{0.40\textwidth}
        \centering
        \includegraphics[width=\textwidth]{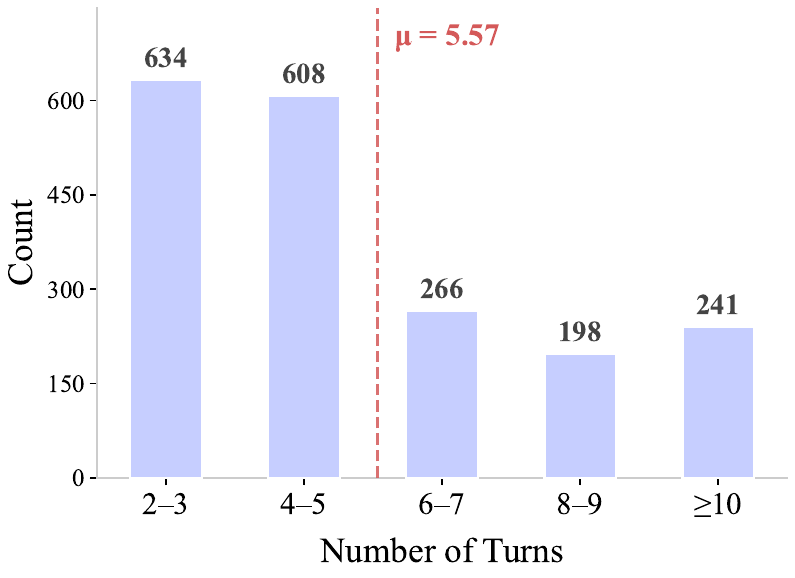}
        \caption{Turn Distribution (NQP-Share)}
        \label{fig:turns_share}
    \end{subfigure}
    \caption{Distribution of conversation lengths. The benchmark retains a significant proportion of long conversations ($\ge 10$ turns), establishing a robust testbed for evaluating memory and cross-turn intent tracking.}
    \label{fig:turns}
\end{figure}

\subsection{Conversation Depth}
To evaluate a model's ability to track long-term intent trajectories, a multi-turn benchmark must preserve sufficient conversational depth. As shown in Figure~\ref{fig:turns}, both NQP-Wild and NQP-Share exhibit a healthy distribution of conversation lengths. Notably, with averages of 4.82 (NQP-Wild) and 5.57 (NQP-Share) turns, the datasets maintain a substantial ``long-tail'' of deep conversations ($\ge 10$ turns). This long-tail volume provides the crucial data foundation for the long-context performance analysis discussed in \S\ref{sec:turn_analysis}.

\subsection{Dynamics of Intent Transfer}
Users in multi-turn dialogues rarely follow a static path. We categorize the dynamic evolution of user intent into four paradigms, defined by their core behavioral characteristics:

\begin{itemize}
    \setlength{\itemsep}{0pt}
    \item \textbf{Deepening:} Focuses on conceptual comprehension, where the user stays on the same core topic but goes deeper to seek a more sophisticated understanding.
    \item \textbf{Application:} Operates at the practical execution level, where the user performs further operations on the AI's previous output to produce a derived artifact.
    \item \textbf{Associated Shift:} Represents logically inferable topic transitions, where the user switches to a different but contextually connected topic.
    \item \textbf{Challenge:} Centers on contradiction and conflict resolution, where the user pushes back on the AI's output to correct substantive errors or flawed assumptions.
\end{itemize}

\begin{figure}[htbp] 
    \centering
    \includegraphics[width=0.74\linewidth]{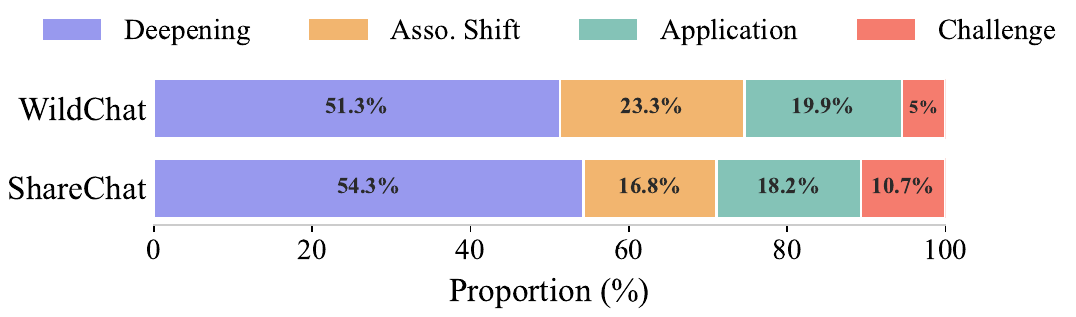} 
    \caption{Proportion of intent transfer paradigms. Over 40\% of the transitions involve cognitive leaps (Application or Associated Shift) rather than simple topic continuation, highlighting the dynamic nature of user intents.}
    \label{fig:intent_transfer}
\end{figure}

Figure~\ref{fig:intent_transfer} reveals that while \textit{Deepening} accounts for over half of the transitions (aligning with the natural human tendency to drill down into a topic), nearly 40\% of the queries involve \textit{Application} or \textit{Associated Shift}. This confirms that accurate next-query prediction requires the model to anticipate dynamic intent evolutions rather than simply extending the current context.

\subsection{Difficulty}
To ensure the benchmark possesses sufficient headroom to distinguish advanced models, we classify the cognitive difficulty of each prediction target using a three-dimensional gating framework. A prediction is classified as \textbf{Hard} if it triggers any of the following conditions:
\begin{itemize}
    \setlength{\itemsep}{0pt}
    \item \textbf{Contextual Distance:} The prediction requires information from $\ge 2$ turns ago (i.e., it cannot be predicted using only the latest exchange); for \textit{Associated Shift}, the polarity is reversed because a distant anchor makes the shift more predictable, whereas a jump without prior grounding is inherently harder.
    \item \textbf{Predictive Entropy:} The current dialogue state presents $\ge 5$ independent and equiprobable future directions.
    \item \textbf{Reasoning Gap:} Establishing the connection between the dialogue history and the target query requires implicit reasoning or external domain knowledge.
\end{itemize}

\begin{figure}[htbp]
    \centering
    \includegraphics[width=0.74\linewidth]{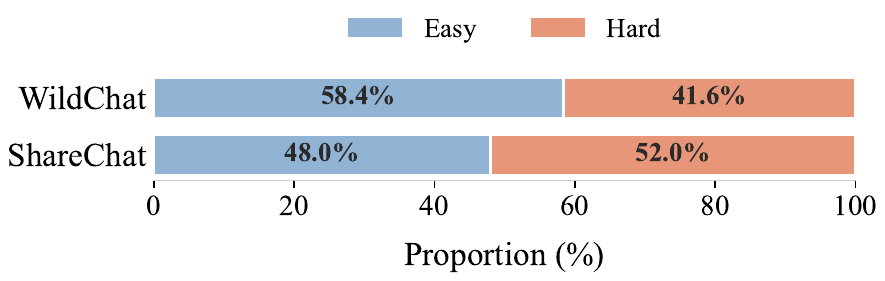} 
    \caption{Distribution of sample difficulty. Nearly half of the benchmark consists of \textit{Hard} instances, requiring long-context dependency, disambiguation of high-entropy states, or implicit reasoning.}
    \label{fig:difficulty}
\end{figure}

As depicted in Figure~\ref{fig:difficulty}, under this rigorous non-linear criterion, 41.6\% of NQP-Wild and 52.0\% of NQP-Share samples are categorized as \textit{Hard}. This substantial proportion of challenging instances prevents models from achieving high scores via superficial pattern matching, ensuring a rigorous evaluation of true agentic reasoning.

\section{Curation Pipeline Attrition}
\label{app:attrition}

Table~\ref{tab:attrition} reports the number of samples retained at each pipeline stage for the two public subsets.

\begin{table}[ht]
\centering
\small
\renewcommand{\arraystretch}{1.10}
\setlength{\tabcolsep}{4pt}
\begin{tabular}{lrcrc}
\toprule
\multirow{2}{*}{Stage} & \multicolumn{2}{c}{NQP-Wild} & \multicolumn{2}{c}{NQP-Share} \\
\cmidrule(lr){2-3} \cmidrule(lr){4-5}
& \#Samples & Ret.\% & \#Samples & Ret.\% \\
\midrule
Raw corpus          & 3{,}199K & --   & 130K   & -- \\
Stage I (heuristic) & 148K     & 4.6  & 30.5K  & 23.6 \\
Stage II (GLM-5)    & 29.2K    & 19.7 & 6.1K   & 20.0 \\
Stage III (Gemini)  & 21.0K    & 72.0 & 4.7K   & 77.2 \\
+ Trunc.\ mining    & +6.7K    & --   & +1.2K  & -- \\
Final               & 18.3K    & --   & 1.9K   & -- \\
\bottomrule
\end{tabular}
\caption{
Filtering attrition for the two public NQP-Bench subsets. ``Ret.\%'' is the retention rate relative to the previous stage. Truncation mining recovers predictable prefixes from the Stage~II DROP pool.
}
\label{tab:attrition}
\end{table}

Stage~I attrition in WildChat (95.4\%) is dominated by language filtering (the corpus is multilingual; only English is retained). Stage~II applies the most aggressive filter: ${\sim}80\%$ of heuristically clean samples are judged unpredictable, reflecting the inherent difficulty of the task. Stage~III confirms 72--77\% of forwarded samples, overturning roughly one quarter upon expert review.
After pooling Stage~III KEEP and truncation-mined samples, a final quality pass removes near-duplicates (by embedding similarity) and over-length samples, reducing the pool to the released benchmark size, followed by a stratified train/test split balance of difficulty (Appendix~\ref{app:dataset_statistics}).

\section{Case Study}
\label{app:case_study}

\begin{figure*}[t]
\centering
\resizebox{0.86\textwidth}{!}{%
\begin{tikzpicture}[
    card/.style={draw=gray!40, rounded corners=2pt, fill=white,
        inner sep=4pt, text width=5.5cm, align=left, font=\small},
    mem/.style={draw=gray!50, rounded corners=2pt, fill=gray!4,
        inner sep=5pt, text width=5.5cm, align=left, font=\small},
    chip/.style={rounded corners=2pt, inner sep=2pt, font=\footnotesize\bfseries,
        minimum width=1.1cm, minimum height=0.45cm, align=center},
    arr/.style={->, >=stealth, gray!60, line width=0.7pt},
    darr/.style={->, >=stealth, gray!50, line width=0.6pt},
]

\node[font=\normalsize\bfseries, text=black!80] at (3.5, 0.6) {Conversation History};
\node[font=\scriptsize\itshape, text=gray] at (3.5, 0.1) {(user queries abbreviated; 13 turns total)};
\node[font=\normalsize\bfseries, text=black!80] at (13, 0.6) {Memory Evolution};
\node[font=\scriptsize\itshape, text=gray] at (13, 0.1) {(4 of 13 snapshots shown; excerpted)};

\node[card] (c1) at (3.5, -1.5) {
    {\color{gray}\scriptsize\bfseries Turns 1--2}\\[1pt]
    Top 10 plants highest in protein\\
    Ranking errors in protein list
};
\draw[left color=blue!60, right color=blue!60, draw=none, rounded corners=1pt]
    ([xshift=-0.5pt]c1.north west) rectangle ([xshift=2.5pt]c1.south west);
\node[chip, fill=blue!10, draw=blue!50, text=blue!70!black] at (-0.3, -1.5) {Nutrition};

\node[mem] (m1) at (13, -1.5) {
    {\bfseries\footnotesize Memory after Turn 2}\\[2pt]
    {\itshape\color{gray!80!black}
    User interested in nutritional information.\\
    Requested high-protein plant list.\\
    Wants data accuracy and comparison.\\[-1pt]
    \color{gray!50}\hfill[\,\dots\,]}
};
\draw[arr] (c1.east) -- (m1.west);

\node[card] (c2a) at (3.5, -3.5) {
    {\color{gray}\scriptsize\bfseries Turns 3--5}\\[1pt]
    Nutritional protein data corrections\\
    Recurring numbness symptoms
};
\draw[left color=orange!70, right color=orange!70, draw=none, rounded corners=1pt]
    ([xshift=-0.5pt]c2a.north west) rectangle ([xshift=2.5pt]c2a.south west);
\node[chip, fill=orange!10, draw=orange!50, text=orange!70!black] at (-0.3, -3.5) {Health};

\node[card] (c2b) at (3.5, -5.0) {
    {\color{gray}\scriptsize\bfseries Turns 6--8}\\[1pt]
    Foods high in chlorogenic acid\\
    Back to nutrition research
};
\draw[left color=blue!60, right color=blue!60, draw=none, rounded corners=1pt]
    ([xshift=-0.5pt]c2b.north west) rectangle ([xshift=2.5pt]c2b.south west);
\node[chip, fill=blue!10, draw=blue!50, text=blue!70!black] at (-0.3, -5.0) {Nutrition};

\node[mem] (m2) at (13, -4.25) {
    {\bfseries\footnotesize Memory after Turn 8}\\[2pt]
    {\itshape\color{gray!80!black}
    TOPIC: nutrition $\to$ health $\to$ nutrition.\\
    Tracking chlorogenic acid data.\\
    Prefers precise nutritional information.\\[-1pt]
    \color{gray!50}\hfill[\,\dots\,]}
};
\coordinate (j2) at (7.2, -4.25);
\draw[gray!60, line width=0.7pt] (c2a.east) -| (j2);
\draw[gray!60, line width=0.7pt] (c2b.east) -| (j2);
\draw[arr] (j2) -- (m2.west);

\node[card] (c3) at (3.5, -7.0) {
    {\color{gray}\scriptsize\bfseries Turns 9--11}\\[1pt]
    Recipe: mushroom, cabbage, broccoli stir-fry\\
    Applying nutritional knowledge to cooking
};
\draw[left color=green!60, right color=green!60, draw=none, rounded corners=1pt]
    ([xshift=-0.5pt]c3.north west) rectangle ([xshift=2.5pt]c3.south west);
\node[chip, fill=green!10, draw=green!50!black, text=green!50!black] at (-0.3, -7.0) {Cooking};

\node[mem] (m3) at (13, -7.0) {
    {\bfseries\footnotesize Memory after Turn 11}\\[2pt]
    {\itshape\color{gray!80!black}
    TOPIC: nutrition $\to$ health $\to$ cooking.\\
    User applying nutritional knowledge to\\
    cooking. Recipes link to prior interests.\\[-1pt]
    \color{gray!50}\hfill[\,\dots\,]}
};
\draw[arr] (c3.east) -- (m3.west);

\node[card] (c4a) at (3.5, -9.0) {
    {\color{gray}\scriptsize\bfseries Turn 12}\\[1pt]
    What causes darkness under people's eyes?
};
\draw[left color=purple!60, right color=purple!60, draw=none, rounded corners=1pt]
    ([xshift=-0.5pt]c4a.north west) rectangle ([xshift=2.5pt]c4a.south west);
\node[chip, fill=purple!10, draw=purple!50, text=purple!70!black] at (-0.3, -9.0) {Skincare};

\node[card] (c4b) at (3.5, -10.5) {
    {\color{gray}\scriptsize\bfseries Turn 13}\\[1pt]
    What can be done to retain skin collagen?
};
\draw[left color=purple!60, right color=purple!60, draw=none, rounded corners=1pt]
    ([xshift=-0.5pt]c4b.north west) rectangle ([xshift=2.5pt]c4b.south west);
\node[chip, fill=purple!10, draw=purple!50, text=purple!70!black] at (-0.3, -10.5) {Skincare};

\node[mem] (m4) at (13, -9.75) {
    {\bfseries\footnotesize Memory after Turn 13}\\[2pt]
    {\itshape\color{gray!80!black}
    Skin health + nutrition converging.\\
    Previous nutritional research now\\
    connects to collagen/skincare topics.\\[-1pt]
    \color{gray!50}\hfill[\,\dots\,]}
};
\coordinate (j4) at (7.2, -9.75);
\draw[gray!60, line width=0.7pt] (c4a.east) -| (j4);
\draw[gray!60, line width=0.7pt] (c4b.east) -| (j4);
\draw[arr] (j4) -- (m4.west);

\draw[darr] (m1.south) -- (m2.north);
\draw[darr] (m2.south) -- (m3.north);
\draw[darr] (m3.south) -- (m4.north);

\draw[dashed, gray!40] (-1.5, -12.0) -- (16.5, -12.0);
\draw[darr] (8.25, -11.8) -- (8.25, -12.2);

\node[draw=red!60!black, rounded corners=2pt, fill=red!3,
    inner sep=5pt, text width=6cm, align=left, line width=0.8pt, font=\small]
    (pred) at (3.5, -13.2) {
    {\bfseries\color{red!60!black} Predicted (Top Candidate)}\\[2pt]
    {\itshape ``What foods are best for increasing collagen production?''}
};

\node[draw=green!50!black, rounded corners=2pt, fill=green!3,
    inner sep=5pt, text width=6cm, align=left, line width=0.8pt, font=\small]
    (gt) at (13, -13.2) {
    {\bfseries\color{green!40!black} Ground Truth}\\[2pt]
    {\itshape ``Which foods are best to increase collagen production?''}
};

\node[fill=green!50!black, text=white, font=\small\bfseries,
    rounded corners=2pt, inner sep=3pt, align=center]
    at (8.25, -13.2) {Exact\\[-1pt]Match};

\node[font=\footnotesize\itshape, text=gray]
    at (8.25, -14.4) {17.6K chars $\to$ 338 chars\quad(52$\times$ compression)};

\end{tikzpicture}%
}
\caption{Qualitative example from NQP-Wild. Left: abbreviated user queries from a 13-turn conversation spanning four topic clusters (assistant responses omitted). Right: four representative snapshots of the recursive memory (excerpted; the full memory is updated at every turn). The memory compresses 17.6K characters into 338 characters ($52\times$) while preserving the cross-topic intent trajectory, enabling an exact intent match with the ground-truth next query.}
\label{fig:case_study}
\end{figure*}
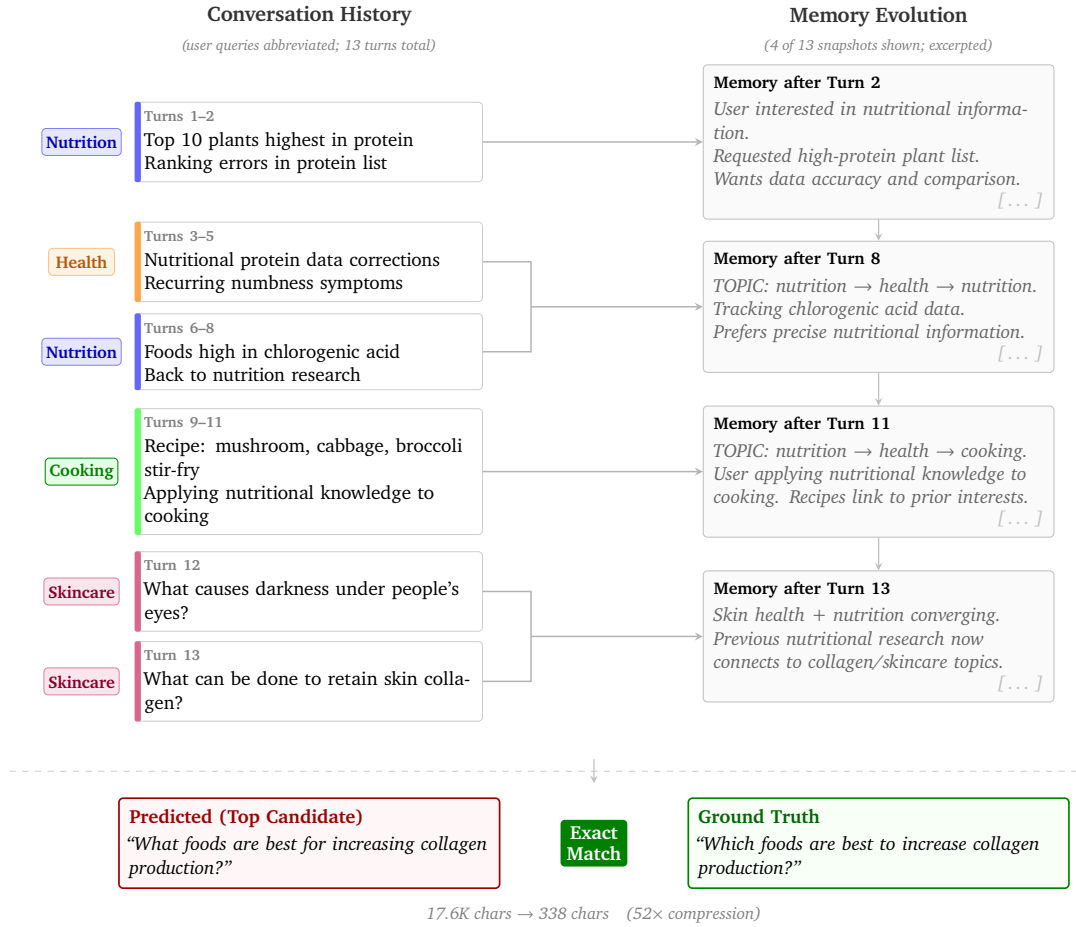

Figure~\ref{fig:case_study} presents a representative 13-turn example from NQP-Wild illustrating how recursive memory supports prediction across extended, topic-shifting conversations.
The user begins with questions about protein content in foods (nutrition), transitions through health conditions, chlorogenic-acid-rich foods, cooking recipes, and finally arrives at skin collagen, spanning four distinct topic clusters over 13 turns.
The ground-truth next query is ``Which foods are best to increase collagen production?'', a question that bridges the user's early nutritional interest (turns~1--11) with the recent collagen focus (turns~12--13).

This cross-topic synthesis exposes the limitations of both baselines.
A \textbf{Current-turn} model sees only the final exchange about skincare routines for collagen retention; without access to the earlier food-and-nutrition thread, it lacks the signal needed to predict a food-oriented follow-up and would instead predict skincare-related questions.
A \textbf{Full-history} model receives all 13 turns ($>$14K tokens), but the predictive signal, the user's persistent interest in food and nutrition, is buried among recipe corrections, a digression on numbness symptoms (turns~3--5), and lengthy assistant responses, making it harder to distill the relevant cross-topic thread.

Our method compresses 17.6K characters of raw conversation into a 338-character memory ($52\times$ compression) that selectively retains the evolving intent trajectory: \emph{food/nutrition $\to$ health $\to$ cooking $\to$ skin-health/collagen}.
From this compact state, the model correctly predicts ``What foods are best for increasing collagen production?'' as its top candidate, achieving an exact intent match with the ground truth.
The example illustrates the core advantage of recursive memory: it discards turn-level noise while preserving the cross-topic predictive signal that neither a single-turn window nor an uncompressed full history can reliably surface.

\makeatletter\setlength{\@fptop}{0pt}\makeatother

\section{Additional Baselines}
\label{app:additional_baselines}

To isolate the effect of history interface design from task-specific training, we compare all five interfaces using Gemini-3.1-Pro in a zero-shot setting on NQP-Wild, ensuring no method benefits from dedicated optimization.

\begin{itemize}[itemsep=0pt,topsep=2pt]
\item \textbf{Sliding-window} ($w{=}3$): The model receives only the most recent $w = 3$ user--assistant exchanges as input, discarding all earlier turns.
\item \textbf{Summarize-then-predict}: At each turn, the model first generates a free-form summary of the full conversation history in a single pass, then predicts the next query conditioned on this summary.
\end{itemize}

\begin{table}[ht]
\centering
\small
\renewcommand{\arraystretch}{1.08}
\begin{tabular*}{0.64\linewidth}{@{\extracolsep{\fill}}lcc}
\toprule
Method & Judge & Human \\
\midrule
Current-turn                & 45.53 & 45.31 \\
Sliding-window ($w{=}3$)    & 46.15 & 46.38 \\
Summarize-then-predict      & 47.42 & 47.80 \\
Full-history                & 48.81 & 49.86 \\
\addlinespace[2pt]
Ours                        & \textbf{50.72} & \textbf{51.14} \\
\bottomrule
\end{tabular*}
\caption{Comparison with additional baselines on NQP-Wild (Gemini-3.1-Pro, zero-shot). All methods use the same model without task-specific training, isolating the effect of history interface design.}
\label{tab:additional_baselines}
\end{table}

As shown in Table~\ref{tab:additional_baselines}, the ranking Ours $>$ Full-history $>$ Summarize-then-predict $>$ Sliding-window $>$ Current-turn holds consistently.
Sliding-window ($w{=}3$) provides only a marginal gain over Current-turn ($+0.62$), confirming that a fixed recent window captures little additional predictive signal beyond the latest exchange.

Summarize-then-predict outperforms Current-turn ($+1.89$), showing that a frontier model can extract useful cross-turn context through summarization. However, it still falls short of Full-history ($-1.39$), indicating that even high-quality task-agnostic compression loses predictive signals that raw history preserves.
Our recursive memory outperforms Full-history by $+1.91$ and Summarize-then-predict by $+3.30$, demonstrating that the advantage stems from the structure of prediction-oriented recursive compression, not merely from having access to cross-turn context.

\section{Hyperparameter Sensitivity}
\label{app:hyperparameter}

We ablate inference-time hyperparameters ($k$ and $N$) using Gemini-3.1-Pro in a zero-shot setting on NQP-Wild, isolating the effect of each parameter from task-specific training.
We separately ablate the training-time reward weight~$\lambda$ using RL-trained Qwen, where each value is independently trained.

\paragraph{Memory budget $k$.}

\begin{table}[ht]
\centering
\small
\renewcommand{\arraystretch}{1.08}
\begin{tabular*}{0.72\linewidth}{@{\extracolsep{\fill}}lccc}
\toprule
$k$ (tokens) & Judge & Human & Avg.\ Tokens/Turn \\
\midrule
100   & 47.80 & 48.20 & ${\sim}$250 \\
200   & 49.35 & 49.70 & ${\sim}$350 \\
500   & \textbf{50.72} & \textbf{51.14} & ${\sim}$650 \\
800   & 50.50 & 50.85 & ${\sim}$950 \\
1{,}000 & 50.25 & 50.60 & ${\sim}$1{,}150 \\
\bottomrule
\end{tabular*}
\caption{Effect of memory budget $k$ on NQP-Wild (Gemini-3.1-Pro, zero-shot).}
\label{tab:ablation_k}
\end{table}

Performance increases steadily from $k{=}100$ to $k{=}500$ as the model gains capacity to retain richer intent trajectories. Beyond $k{=}500$, performance plateaus and slightly declines: with a larger budget, the memory tends to retain verbose turn-level details that dilute predictive signals. The sweet spot at $k{=}500$ balances sufficient capacity with effective information bottlenecking. Even at $k{=}1{,}000$ the per-turn cost (${\sim}1{,}150$ tokens) remains far below Full-history (${\sim}14{,}000$ tokens at 14 turns).

\paragraph{Number of candidate predictions $N$.}

\begin{table}[ht]
\centering
\small
\renewcommand{\arraystretch}{1.08}
\begin{tabular*}{0.58\linewidth}{@{\extracolsep{\fill}}lcc}
\toprule
$N$ & Judge & Human \\
\midrule
1 & 49.15 & 49.80 \\
2 & 50.22 & 50.70 \\
3 & \textbf{50.72} & \textbf{51.14} \\
5 & 50.85 & 51.28 \\
\bottomrule
\end{tabular*}
\caption{Effect of the number of candidate predictions $N$ on NQP-Wild (Gemini-3.1-Pro, zero-shot).}
\label{tab:ablation_n}
\end{table}

$N{=}1$ incurs a notable penalty ($-1.57$), as a single prediction must cover all ambiguity in the user's next intent. The gain from $N{=}2$ to $N{=}3$ is moderate ($+0.50$), and $N{=}5$ offers only marginal further improvement ($+0.13$), indicating that three candidates suffice to cover the main intent hypotheses without wasting generation budget.

\paragraph{Reward weight $\lambda$.}
Since $\lambda$ is a training-time parameter, each value below corresponds to an independently trained model (RL-trained Qwen on NQP-Wild).

\begin{table}[ht]
\centering
\small
\renewcommand{\arraystretch}{1.08}
\begin{tabular*}{0.58\linewidth}{@{\extracolsep{\fill}}lcc}
\toprule
$\lambda$ & Judge & Human \\
\midrule
0.5 & 44.85 & 45.60 \\
0.7 & 45.48 & 46.30 \\
0.9 & \textbf{46.00} & \textbf{47.27} \\
1.0 & 45.62 & 46.55 \\
\bottomrule
\end{tabular*}
\caption{Effect of reward weight $\lambda$ on NQP-Wild (RL-trained Qwen).}
\label{tab:ablation_lambda}
\end{table}

When $\lambda$ is too low ($0.5$), the format penalty dominates and the model under-optimizes for prediction quality. At $\lambda{=}1.0$ (pure judge reward, no format penalty), performance drops slightly because occasional malformed outputs corrupt the multi-turn agent loop during training. $\lambda{=}0.9$ strikes the best balance, allocating most of the reward signal to prediction quality while maintaining sufficient format compliance to keep training stable.

\section{Statistical Significance}
\label{app:significance}

To verify that our improvements are statistically reliable, we report bootstrap confidence intervals for the main comparison on NQP-Wild (RL-trained Qwen).
We resample the test set ($n{=}2{,}354$) 1{,}000 times with replacement and compute the LLM Judge score for each bootstrap sample.
We further compute paired bootstrap $p$-values by counting the fraction of resamples in which the baseline score exceeds ours.

\begin{table}[ht]
\centering
\small
\renewcommand{\arraystretch}{1.08}
\begin{tabular*}{0.78\linewidth}{@{\extracolsep{\fill}}lcc}
\toprule
Method & Judge (mean $\pm$ 95\% CI) & $p$-value vs.\ Ours \\
\midrule
Current-turn  & $44.02 \pm 0.78$ & $< 0.001$ \\
Full-history  & $44.35 \pm 0.82$ & $< 0.01$ \\
Ours          & $46.00 \pm 0.75$ & --- \\
\bottomrule
\end{tabular*}
\caption{Bootstrap confidence intervals and paired bootstrap $p$-values on NQP-Wild (RL-trained Qwen, 1{,}000 resamples).}
\label{tab:significance}
\end{table}

The $95\%$ confidence intervals of our method do not overlap with those of either baseline, and both paired $p$-values are well below $0.01$, confirming that the improvements are statistically significant.

\section{Judge Validation}
\label{app:judge_validation}

To verify that the LLM-judge protocol is a reliable proxy for human judgment, we analyze agreement at three levels using the 250 human-annotated samples per subset.

\paragraph{Human inter-annotator agreement.}
Five expert annotators independently scored each sample on the same 5-point rubric used by the LLM judges.
Fleiss' $\kappa$ across the five annotators is $0.72$ (NQP-Priv), $0.75$ (NQP-Wild), and $0.70$ (NQP-Share), indicating substantial agreement and confirming that the rubric yields consistent human judgments.

\paragraph{Human--LLM correlation.}
We compute Spearman's $\rho$ between the per-sample LLM-judge score (majority vote of three judges) and the corresponding human score (majority vote of five annotators) on the shared 250-sample subsets.
The correlation is $0.78$ (NQP-Priv), $0.81$ (NQP-Wild), and $0.76$ (NQP-Share), showing strong sample-level agreement beyond the ranking-level consistency reported in \S\ref{sec:main_results}.

\paragraph{LLM inter-judge agreement.}
Fleiss' $\kappa$ among the three LLM judges (Claude Opus 4.7, GPT-5.5, Gemini-3.1-Pro) is $0.68$ (NQP-Priv), $0.71$ (NQP-Wild), and $0.66$ (NQP-Share). The moderate-to-substantial agreement supports the use of majority vote to aggregate their ratings.

\paragraph{Human validation of benchmark curation.}
Since the curation pipeline relies on LLM judgments to assess predictability, we validate that retained samples reflect genuine predictability rather than LLM-specific biases. Three human annotators independently reviewed 400 randomly sampled instances from the curation candidate pool (200 retained KEEP instances and 200 rejected DROP instances) and judged whether the target query was predictable from the conversation context. Human--LLM agreement on the KEEP/DROP decision reached $92\%$ (Cohen's $\kappa = 0.83$), confirming that the LLM cascade reliably identifies predictable conversations. Furthermore, the evaluation judges (Claude, GPT-5.5, Gemini-3.1-Pro) are largely distinct from the primary curation model (GLM-5), and Gemini serves only as a secondary reviewer in Stage~III, limiting the overlap between data selection and scoring.

\section{Score Distribution Analysis}
\label{app:score_dist}

Table~\ref{tab:score_dist} shows the full rubric-level score distribution on NQP-Wild using the recursive memory interface, comparing an untrained base model against the RL-trained model and a frontier API.

\begin{table}[ht]
\centering
\small
\renewcommand{\arraystretch}{1.08}
\begin{tabular*}{0.74\linewidth}{@{\extracolsep{\fill}}lcccccc}
\toprule
Model & 1 & 2 & 3 & 4 & 5 & Mean \\
\midrule
Base Qwen       & 26 & 24 & 25 & 16 &  9 & 39.57 \\
RL Qwen         & 21 & 22 & 23 & 20 & 14 & 46.00 \\
Gemini-3.1-Pro  & 15 & 22 & 25 & 22 & 16 & 50.72 \\
\bottomrule
\end{tabular*}
\caption{Score distribution (\%, rounded) by rubric level on NQP-Wild. All rows use recursive memory. 1\,=\,Irrelevant, 2\,=\,Slightly related, 3\,=\,Topic related, 4\,=\,Highly aligned, 5\,=\,Intent hit.}
\label{tab:score_dist}
\end{table}

RL training shifts the distribution markedly toward higher rubric levels. Compared to the untrained base, score~$\geq 4$ predictions (Highly aligned or Intent hit) rise from $25\%$ to $34\%$ ($+9$ pp), while score~$\leq 2$ predictions drop from $50\%$ to $43\%$ ($-7$ pp). Gemini-3.1-Pro extends this trend further, reaching $38\%$ at score~$\geq 4$ and only $37\%$ at score~$\leq 2$. The moderate overall mean ($\sim\!50$) does not reflect uniformly mediocre predictions; rather, a substantial fraction of predictions are actionable (score $\geq 4$), and the mean is pulled down by genuinely difficult samples that remain in the lower tiers. We note that open-ended next-query prediction is inherently harder than constrained tasks such as next-utterance selection or slot filling, because any turn in a multi-turn conversation may branch into multiple plausible directions. The $41\%$--$52\%$ of Hard samples in NQP-Bench (Figure~\ref{fig:difficulty}) further depress absolute scores. Within this challenging setting, the consistent relative gains of RL training ($+6.4$ over Base) and our method over Full-history demonstrate meaningful progress.

From a deployment perspective, the $34\text{--}38\%$ of predictions reaching score $\geq 4$ are directly useful for downstream applications such as follow-up suggestion, pre-fetching retrieval results, and speculative response generation. Even score-$3$ predictions, which capture the correct topic but miss the exact question, can inform coarse-grained pre-computation (e.g., warming a relevant document cache). Because the predictor runs asynchronously while the user is still reading the current response, even partially correct predictions incur no user-facing latency cost and can be silently discarded when inaccurate.

\section{Fine-Grained Evaluation and Error Analysis}
\label{app:failure}

To gain a comprehensive understanding of our method, we first stratify performance by difficulty and intent transfer paradigm (Table~\ref{tab:cross_analysis}), then analyze the dominant error patterns.

\paragraph{Fine-grained performance breakdown.}
Aggregating across difficulty levels, our method outperforms Full-history on three of the four paradigms: Deepening ($+3.9$), Application ($+2.1$), and Challenge ($+3.0$), demonstrating broad effectiveness across diverse intent dynamics.
The largest cell-level gains appear on Easy $\times$ Deepening ($+5.1$) and Hard $\times$ Challenge ($+5.7$), where the intent chain naturally tracks progressive exploration or unresolved tensions in the assistant's output.
The sole exception is Associated Shift, where Full-history leads by $+3.3$ on Easy samples.
This is expected: when the next query connects to non-adjacent earlier details through a grounded lateral association, the memory may have already compressed away the relevant earlier context, while raw history preserves these latent cross-topic links.

\begin{table}[ht]
\centering
\small
\renewcommand{\arraystretch}{1.08}
\begin{tabular*}{0.80\linewidth}{@{\extracolsep{\fill}}ll cccc}
\toprule
\multirow{2}{*}{Difficulty} & \multirow{2}{*}{Method} & \multicolumn{4}{c}{Intent Transfer Paradigm} \\
\cmidrule(lr){3-6}
& & Deep. & Appl. & Assoc. & Chall. \\
\midrule
\multirow{2}{*}{Easy}
& Full-hist   & 45.1 & 42.0 & \textbf{50.1} & 44.3 \\
& Ours        & \textbf{50.2} & \textbf{44.6} & 46.8 & \textbf{45.1} \\
\addlinespace[3pt]
\multirow{2}{*}{Hard}
& Full-hist   & 44.6 & 38.8 & \textbf{44.5} & 39.5 \\
& Ours        & \textbf{45.8} & \textbf{40.1} & 42.7 & \textbf{45.2} \\
\bottomrule
\end{tabular*}
\caption{Performance (LLM Judge $\times 100$) stratified by difficulty and intent transfer paradigm on NQP-Wild (RL-trained Qwen). Bold indicates the better method.}
\label{tab:cross_analysis}
\end{table}

\paragraph{Detail loss under compression.}
The bounded memory is optimized to retain high-level intent trajectories, but fine-grained details (e.g., specific numbers, exact constraints, or minor sub-topics) may be compressed away if the model judges them less salient at the time of the update. When the user's next query happens to depend on such details, prediction quality suffers. This pattern is most pronounced on Easy samples with Associated Shift, where Full-history outperforms our method by $+3.3$ ($50.1$ vs.\ $46.8$ in Table~\ref{tab:cross_analysis}), because the relevant detail often originates from a non-adjacent earlier turn that the memory has already summarized away.

\paragraph{Error distribution.}
Across all Hard samples on NQP-Wild, our method produces scores $\leq 2$ (Slightly related or Irrelevant) on approximately $18\%$ of instances.
Qualitative inspection reveals two dominant patterns:
(1)~\emph{detail-loss errors}, where the memory captures the overall intent trajectory but drops a specific detail that the next query depends on; and
(2)~\emph{ambiguity errors}, where the conversation state genuinely supports multiple plausible next queries and the model's top candidates diverge from the ground truth despite being contextually reasonable.

\end{document}